\def\paperlanguage{}
\pdfoutput=1 

\documentclass[letterpaper, 10 pt, conference]{ieeeconf}  

\usepackage{bm}
\usepackage{cite}
\include{preamble}

\IEEEoverridecommandlockouts                              

\title{\LARGE \textbf
  {
    \switchlanguage%
    {%
      Designing Fluid-Exuding Cartilage for Biomimetic Robots \\ Mimicking Human Joint Lubrication Function
    }%
    {%
      人の関節潤滑機能を模した人体模倣ロボットの \\ 液体滲出軟骨機構の構成法
    }%
  }
}

\author{
  Akihiro Miki, Yuta Sahara, Kazuhiro Miyama, Shunnosuke Yoshimura, Yoshimoto Ribayashi, \\
  Shun Hasegawa, Kento Kawaharazuka, Kei Okada, and Masayuki Inaba
  \thanks{The authors are with the Department of Mechano-Informatics, Graduate School of Information Science and Technology, The University of Tokyo, 7-3-1 Hongo, Bunkyo-ku, Tokyo, 113-8656, Japan.
    {\texttt\small [miki, sahara, miyama, yoshimura, ribayashi, hasegawa, kawaharazuka, k-okada, inaba]@jsk.t.u-tokyo.ac.jp}
  }
}
\begin{document}

\maketitle
\thispagestyle{empty}
\pagestyle{empty}

%
\begin{abstract}
  \switchlanguage%
  {%
    The human joint is an open-type joint composed of bones, cartilage, ligaments, synovial fluid, and joint capsule, having advantages of flexibility and impact resistance.
    However, replicating this structure in robots introduces friction challenges due to the absence of bearings.
    To address this, our study focuses on mimicking the fluid-exuding function of human cartilage.
    We employ a rubber-based 3D printing technique combined with absorbent materials to create a versatile and easily designed cartilage sheet for biomimetic robots.
    We evaluate both the fluid-exuding function and friction coefficient of the fabricated flat cartilage sheet.
    Furthermore, we practically create a piece of curved cartilage and an open-type biomimetic ball joint in combination with bones, ligaments, synovial fluid, and joint capsule
    to demonstrate the utility of the proposed cartilage sheet in the construction of such joints.
  }%
  {%
    人間の関節は，骨，軟骨，靭帯，滑液，および関節包から成る開放型の関節であり，柔軟性と耐衝撃性の利点を有しています．
    ただし，この構造をロボットに複製すると，軸受がないことから摩擦の問題が発生します．
    このため，私たちの研究では，人間の軟骨の液体滲出機能を模倣することに焦点を当てています．
    ゴムベースの3Dプリント技術と吸水性素材を組み合わせて，生物模倣ロボットのために汎用性が高く設計が容易な軟骨シートを作成します．
    製造した平面状の軟骨シートで，液体滲出機能と摩擦係数の両方を評価します．
    さらに，曲線状の軟骨を作成し，骨，靭帯，滑液，および関節包と組み合わせて，提案した軟骨シートがそのような関節の構築に有用であることを実証します．
  }%
\end{abstract}

\section{Introduction}
\label{sec:introduction}
\switchlanguage%
{%


\begin{figure}[htbp]
  \begin{center}
    \includegraphics[width=\linewidth]{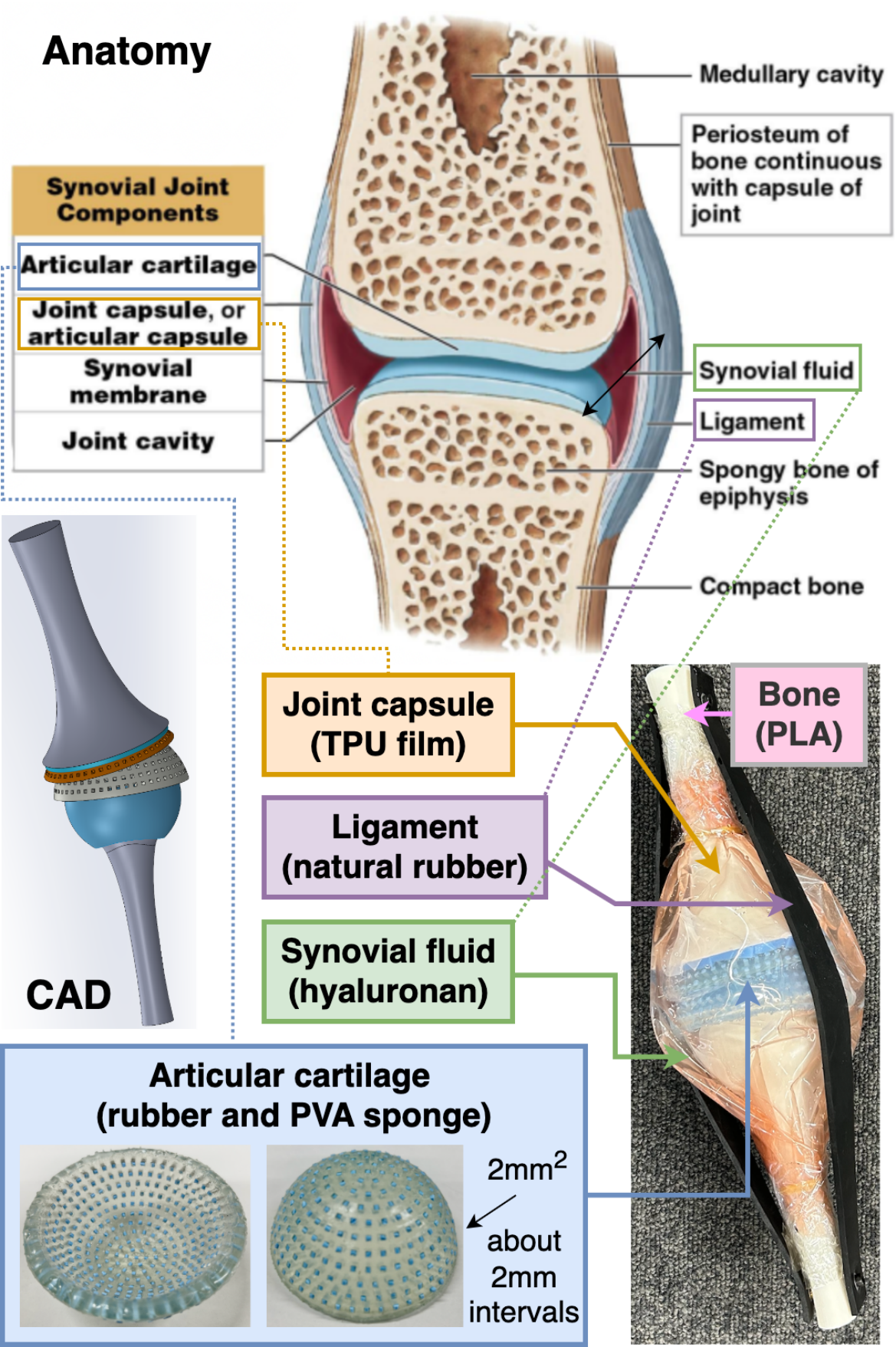}
    \caption{Overview of an open-type ball joint utilizing fluid-exuding cartilage with anatomical figure (adapted from Martini et al., 2017\cite{martini2018anatomy}).}
    \label{fig:open-type-ball-joint-overview}
  \end{center}
\end{figure}
  Currently, robots typically employ axis-driven mechanisms, where the links of each body part are interconnected through joints comprising bearings and gears.
  This established structural design has found widespread application in industrial robots employed within manufacturing facilities.

  In contrast, human joints exhibit open-type ball joints, where bones are interconnected by soft tissues like muscles, ligaments, and joint capsules.
  Human joints possess characteristics such as softness, lightweight, shock resistance, and adaptability to their surroundings due to their soft constraints.
  However, conventional robots lack these inherent properties.

  To enable the widespread integration of robots into human living spaces in the future, it is important that they possess these flexible physical properties.
  Biomimetic robots have been developed to leverage the physical advantages exhibited by living organisms.
  For instance, Kengoro\cite{asano2017design}, a humanoid robot, mimics human tendons based on anatomical knowledge, employing a string-winding mechanism to replicate muscle movements.
  Nevertheless, biomimicry remains an evolving field, and replicating human joint structures poses a considerable challenge due to the complexities associated with managing their flexibility and the incomplete understanding of living organisms.

  A significant hurdle in imitating human joints is the issue of friction.
  Although open-type ball joints have various advantages as described above, they do not use bearings, so simply imitating an open structure would result in too much friction during driving.
  Notably, human joints overcome this problem, exhibiting low friction comparable to that of bearings despite their open structure.
  It has been reported that the low friction of human joints is realized by a lubrication function using a fluid, and the details of the lubrication function have been discussed in various studies\cite{charnley1960lubrication, dowson1966paper, fein1966research, mccutchen1962weeping-lubrication, walker1968boosted, mow1980biphasic}.

  The common point among recent theories is that this lubrication function requires not only the presence of synovial fluid within the joint capsule, but also the presence of fluid-exuding cartilage.
  This cartilage facilitates the release of synovial fluid in response to loads, ensuring a constant presence of synovial fluid between bones even under high loads, consequently contributing to improved lubrication.

  This study aims to mimic the fluid-exuding cartilage and its low-friction properties in the human body to realize an open-type ball joint for biomimetic robots.
  The research involves a preliminary assessment using a prototype planar sheet replicating the function of fluid-exuding cartilage, followed by the creation of spherical cartilage.
  This is subsequently combined with bones, ligaments, and joint capsules to demonstrate the successful construction of an open-type ball joint (\figref{fig:open-type-ball-joint-overview}).

}%
{%
現在のロボットは一般的に，ベアリングやギアによって構成された関節を通して身体各部であるリンクが結合する，軸駆動型ロボットである．
工場で用いる産業用ロボットをはじめとして，広く適用されてきた実績のある身体構造である．
\textcolor{red}{
対して，人間の関節は骨と骨を筋肉，靭帯，関節包といった軟部組織によって柔らかく結ばれた開放型の球関節である．
人間の関節は，関節自体の柔らかさ，軽量性，耐衝撃性，柔らかな拘束ゆえの環境への馴染みやすさといった性質を持つが，一般的なロボットはこれらの性質を持ち合わせていない．
}
今後，ロボットが人間の生活空間へと広く展開されていくためには，これらの柔軟な身体性を有することが重要である．
これまで，このような生体が持つ身体の利点を活用するために生体模倣ロボットが発展してきた．
例えば，筋を模倣した紐を巻き取ることで駆動する，解剖学的知見に基づいた人体模倣型腱駆動ヒューマノイド腱悟郎\cite{asano2017design}がある．
しかし，生体模倣はいまだ不完全であり，特に人間の関節構造の模倣については，その柔軟性の扱いの難しさや生体への理解の不十分さゆえに，困難な課題として残されている．


人体関節の模倣における一つの大きな問題として，摩擦の問題があげられる．
開放型球関節は，上記で述べたような様々な利点を持つが，ベアリングを用いていないため，単純に開放型の構造を模倣するだけでは駆動時の摩擦が大きくなりすぎてしまう．
しかし，人体の関節では摩擦の問題は解決されており，開放型であるにもかかわらず，ベアリングに匹敵するほどの低摩擦であることが知られている．
人体関節の低摩擦性は，液体を用いた潤滑機能によって実現されると報告されており，潤滑機能の詳細については，様々な研究で議論されてきた
\cite{charnley1960lubrication}\cite{dowson1966paper}\cite{fein1966research}\cite{mccutchen1962weeping-lubrication}\cite{walker1968boosted}\cite{mow1980biphasic}．
近年の説に共通して言えることは，この潤滑機能にはただ関節包内にある関節液の存在だけではなく，荷重に応じて関節液がしみだすような，液体滲出軟骨の存在が不可欠であるということである．
液体滲出軟骨の存在によって，高荷重下であっても骨と骨の間に常に関節液が存在でき，潤滑性の向上に貢献している．
そこで本研究では，生体模倣ロボットのための開放型球関節の実現のために，人体に倣った液体滲出軟骨と，その低摩擦性を模倣再現する．
液体滲出軟骨の機能を模倣再現したプロトタイプ平面状シートで基礎評価を行った後に，球面状軟骨を作成し，模擬骨，靭帯，関節包らと組み合わせて開放型球関節(\figref{open-type-ball-joint-overview})を実際に構成できることまで示す．

\begin{figure}[t]
  \begin{center}
    \includegraphics[width=\linewidth]{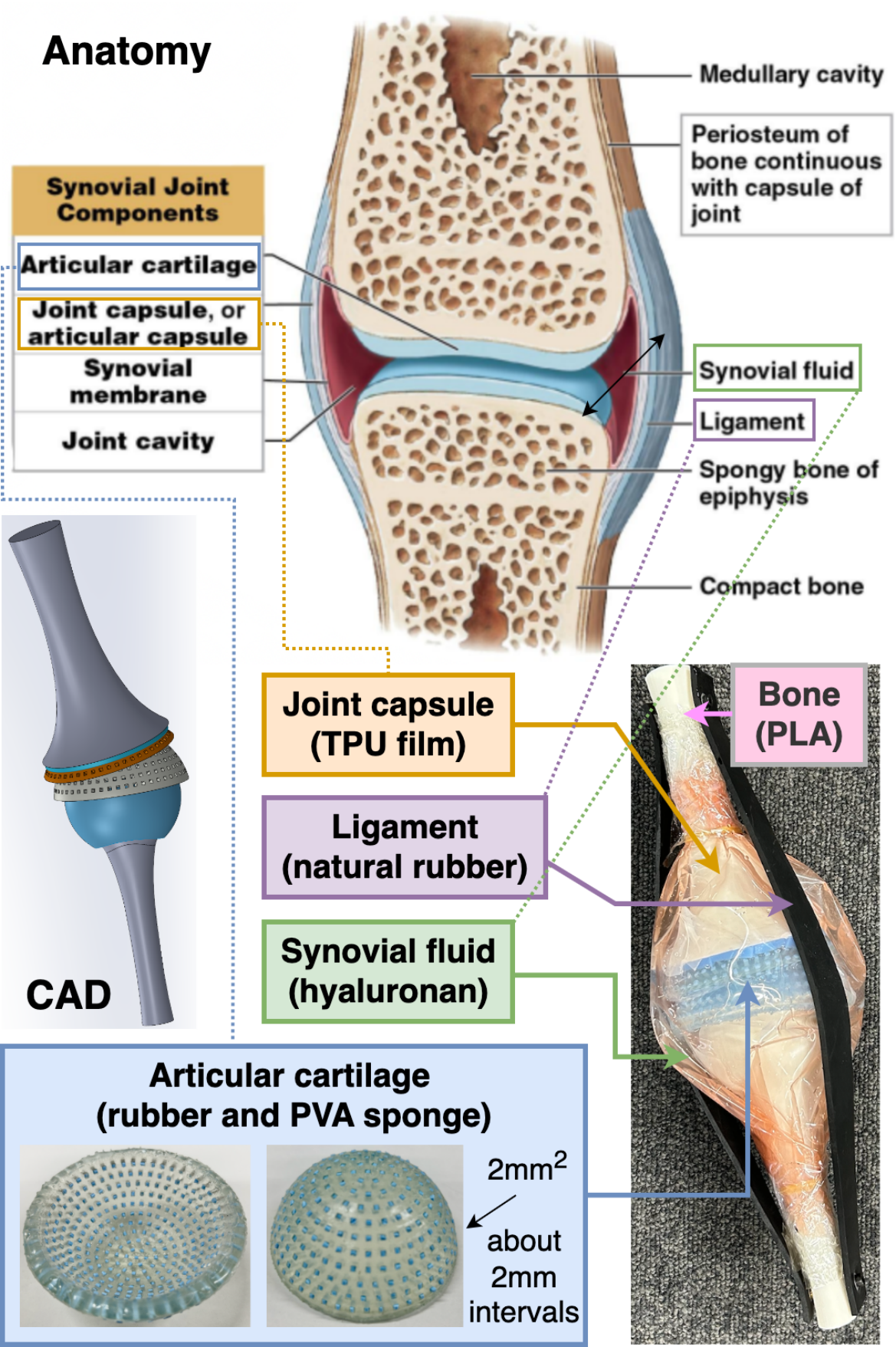}
    \caption{Overview of an open-type ball joint utilizing fluid-exuding cartilage with anatomical figure (adapted from Martini et al., 2017\cite{martini2018anatomy}).}
    \label{fig:open-type-ball-joint-overview}
  \end{center}
\end{figure}
}%

\section{Related Research on Cartilage}
\label{sec:previous_work}
\switchlanguage
{%
  In the field of soft robotics, research on cartilage has been conducted, such as the development of underwater ray-shaped robots made of cartilage\cite{yurugi2021cartilage}.
  However, the applications of cartilage have primarily focused on shaping for soft exterior structures, and there has been little development in the creation of fluid-exuding cartilage for open-type ball joints in life-sized humanoids.
  
  Instead, research on cartilage for joints has been more actively pursued in the medical field rather than in the field of soft robotics.
  In medical research, the development of artificial joints, including cartilage, has progressed.
  For instance, in artificial hip and knee joints, low-friction materials such as cross-linked polyethylene with increased cross-linking structures and its improved version, vitamin E-infused polyethylene, have been utilized in the contact areas of artificial joints\cite{oonishi2000wear}\cite{brooks2002endotoxin}.
  Additionally, low-friction materials in the metal category, such as zirconia ceramics, have been clinically used in the femur\cite{nakamura2002zirconia}.
  However, while these materials exhibit low friction, they lack the fluid lubrication function found in human cartilage.
  As a result, they do not possess the same level of low friction as human cartilage, and they also lack the impact absorption properties provided by the water-containing human cartilage.
  Even if physicians are instructed to avoid strenuous exercise and such use is adhered to, replacement will eventually be necessary.
  Furthermore, polyethylene wear particles can trigger biological reactions, leading to joint effusion and cyst formation\cite{kondo2008anthroscopy}\cite{akisue2001paratibial}.

  In recent years, PVA hydrogel, a structure containing moisture, has been anticipated as a new material for artificial cartilage\cite{sawae1996lubrication}.
  However, challenges remain in terms of mechanical strength, long-term expansion during use, integration with bones and existing tissues, and human compatibility.
  To address the issue of mechanical strength, high-strength PVA hydrogels have been developed by incorporating fibers during gel molding\cite{sakai2019biomimetic-cartilage}.
  However, their creation is challenging and costly.
  Generally, it is still considered difficult to shape specially treated gels through 3D printing.

  In the context of biomimetic robots, human compatibility is not essential, and there is no need to meet the stringent criteria of the medical field.
  Instead, what is more critical is the realization of low friction in open-type ball joints at a low cost and the ability to freely shape the structure through 3D printing according to the robot's design.
  Therefore, in this study, we organize the properties of cartilage, the imitated subject in \secref{sec:cartilage-knowledge}, propose a fluid-exuding cartilage mechanism suitable for biomimetic robots in \secref{sec:cartilage-mechanism}, and evaluate it in \secref{sec:pre-experiment} and beyond.
}%
{%
ソフトロボティクスの分野では，軟骨と軟組織で作成された水中エイ型ロボットの開発などで軟骨の研究が行われてきた\cite{yurugi2021cartilage}．
しかし，軟骨の使用用途は外形を柔らかく形作るためのものであり，等身大ヒューマノイドの開放型球関節を想定しての液体滲出軟骨の開発は行われてこなかった．

むしろ関節に用いる軟骨に関しては，ソフトロボティクス分野よりも医療分野において，活発に研究されてきた．
医療分野では，軟骨を含む人工関節全体の開発が進められてきた．
例えば人工股関節や膝関節において，架橋構造を増やしたクロスリンクポリエチレンや，その改良型であるビタミンE添加型ポリエチレンといった低摩擦素材が人工関節の接触部に用いられてきた\cite{oonishi2000wear}\cite{brooks2002endotoxin}． 
他にも大腿骨ではジルコニアセラミックといった金属系統の低摩擦素材も臨床使用されている\cite{nakamura2002zirconia}．
しかしながら，これらは低摩擦素材ではあるものの，液体潤滑機能を備えてはいない．
そのため，人体の軟骨ほど低摩擦でもなく，水分を含んだ人体軟骨が備えるような衝撃吸収性も乏しい．
医師から激しい運動を避けるように指示され，そのような使用方法を守ってもいずれ交換が必要となる．
また，ポリエチレン摩耗粉が生体反応を引き起こし，関節水腫や嚢胞を形成することもある\cite{kondo2008anthroscopy}\cite{akisue2001paratibial}. 

近年では，水分を含有する構造であるPVAハイドロゲルも人工軟骨の新素材として期待されている\cite{sawae1996lubrication}．
しかし，機械的強度，長期利用における膨張，骨や既存組織との統合，人体適合性に課題が残っている．
機械的強度の課題を解決するために，ゲルを型成形する際に繊維ファイバーを配合して作成する，高強度PVAハイドロゲルも開発されている\cite{sakai2019biomimetic-cartilage}．
しかし，その作成は困難でコストがかかる．
一般的に，特殊処理を施したゲルは3Dプリンティングで手軽に成形することは依然として難しいとされる．

生体模倣ロボットにおいては，人体適合性は不必要であるため，医療分野における厳しい基準を満たす必要はない．
むしろ，低コストで開放型球関節の低摩擦性が実現できることや，ロボットの設計に応じて3Dプリンティングによって自由に形状を成形できることのほうが重要である．
そこで本研究では，第3章で模倣対象である軟骨の性質を整理した上で，第4章で生体模倣ロボットに適した液体滲出軟骨機構を提案し，第5章以降でその評価を行う．
}%

\section{Anatomical and Physiological Foundations of Cartilage}
\label{sec:cartilage-knowledge}
\switchlanguage%
{%
  Cartilage is an elastic and smooth connective tissue, characterized by its translucent and non-porous structure.
  Due to the absence of blood vessels within, cartilage is known for its limited healing capability.

  Based on tissue composition, cartilage can be classified into hyaline cartilage, fibro cartilage, and elastic cartilage.
  Hyaline cartilage, found in joints covering bone ends, as well as in structures like the trachea and pharynx, is highly flexible and capable of smooth movement.
  Fibro cartilage, present in joints with limited mobility such as the spine, serves as a cushion between bones.
  It is slightly harder than articular cartilage, providing firm joint support and shock absorption.
  Elastic cartilage, composed of elastic fibers, has flexibility and is found in structures like the ears.
  This study focuses on articular cartilage, a type of hyaline cartilage.
  In this paper, the term 'cartilage' refers to articular cartilage.

  Articular cartilage consists of an extracellular matrix composed of collagen fibers and proteoglycans, as well as chondrocytes.
  The proteoglycans bind a significant amount of water molecules within their structure, resulting in articular cartilage being a soft tissue containing 70-80\% water.
  The thickness is a few millimeters, and previous studies have measured cartilage thickness on the articular surface of the patella ranging from 0.9-\SI{6.6}{mm}\cite{kladny1996cartilage-measurement}.
  The Young's modulus is reported to be $1.03\pm\SI{0.48}{MPa}$\cite{kabir2020cartilage-protocol}, and the friction coefficient is reported to be between 0.005-0.1\cite{roberts1982modes}.

  Considering that the typical coefficient of friction for solid contact is around 0.3, the reported friction coefficient of articular cartilage is remarkably low.
  Compared to the friction coefficient of rolling friction in bearings (0.0010-0.0015) and sliding bearing friction (0.07-0.15), the friction coefficient of biological cartilage is higher than that of bearings but lower than that of sliding bearings.
  The remarkable low friction of cartilage is supported by the lubrication function of cartilage fluid exudation.
  Under load, fluid exudes from the moisture-rich articular cartilage, fulfilling the lubrication role on the cartilage surface.
  Fluid exudation realizes low friction, protecting the cartilage from the impact forces and wear during joint movement.
  Fluid exudation is load-dependent, ensuring the constant presence of fluid between solid surfaces.
  Additionally, articular cartilage, under high loads, expands the load-bearing area by taking advantage of its low elastic modulus and deformability, thereby reducing contact pressure to a few MPa.
  With these properties, biological cartilage, despite its limited healing capacity, efficiently and durably serves its purpose in the body with low friction.
  Leveraging these characteristics of biological cartilage is valuable in creating open-type ball joints for robots.
}%
{%
軟骨は弾力性があり，なめらかな結合組織の一種である．
半透明で非多孔質な組織である．
内部に血管が侵入しないため，治癒力に乏しいことも知られている．

組織構成によって，硝子軟骨，繊維軟骨，弾性軟骨に大別される．
硝子軟骨は，柔軟性が高く，滑らかに動くことができる，
骨端部分を覆う関節や，気管，咽頭などに存在する．
繊維軟骨は，脊椎など動きの少ない関節に存在し，骨と骨の間でクッションの役割を果たす．
関節軟骨と比べるとやや硬く，関節をしっかりと支え，衝撃を和らげる．
弾性軟骨は弾性繊維から構成されて弾力を持ち，耳などを形作っている．
本研究では硝子軟骨の一種である関節軟骨について扱う．
本論文で用いる軟骨という単語は関節軟骨のことを指す．

関節軟骨はコラーゲン繊維とプロテオグリカンから構成される細胞外マトリックスおよび軟骨細胞から構成されている．
プロテオグリカンが分子内で多量の水分子と結合することで，関節軟骨は70$\sim$80\%の水分を含有する軟組織となっている．
軟骨の厚さは数mm程度であることが知られ，実際に先行研究では0.9$\sim$\SI{6.6}{mm}の軟骨厚さが膝蓋骨の関節面において計測された\cite{kladny1996cartilage-measurement}．
ヤング率は$1.03\pm\SI{0.48}{MPa}$と報告されており\cite{kabir2020cartilage-protocol}，摩擦係数は0.005$\sim$0.1であることも報告されている\cite{roberts1982modes}．

この摩擦係数の値は固体接触における摩擦係数の値が0.3程度であることを考えると，非常に小さな値である．
転がり摩擦のベアリングの摩擦係数が0.0010$\sim$0.0015であることや，すべり軸受の摩擦係数が0.07$\sim$0.15であることを考慮すると，生体軟骨の摩擦係数はベアリングよりも大きく，すべり軸受よりも小さい程度の値であるといえる．
上記の驚くべき低摩擦性は，軟骨の液体滲出機能によって支えられる．
荷重下では，水分を多く含む関節軟骨から液体がしみだし，軟骨表面の潤滑の役目を果たすのである．
液体滲出機能によって低摩擦性が実現され，関節駆動時に関節表面に生じる衝撃力やすり減りから軟骨が守られている．
液体滲出は荷重量に応じたものであり，固体どうしの間に常に液体が存在できるようになっている．
また関節軟骨は，高荷重下において軟骨表面が圧迫されると，軟骨の低弾性率と軟骨支持部の変形性によって荷重支持面積を広く取ることで，接触面圧を数MPaに抑えて負担を軽減していることも知られる．
これらの性質によって，生体軟骨は治癒力に乏しいながらも，低摩擦で効率的に，かつ長い間体内で使用され続けている．
ロボットにおいて，開放型球関節を作成する際には，上記の生体軟骨の特性を活用することが有用である．
}%

\section{Mechanism of Fluid Exudation in Cartilage}
\label{sec:cartilage-mechanism}
\switchlanguage%
{%
  In this section, we propose a mechanism to mimic and replicate the characteristics of biological cartilage discussed in \secref{sec:cartilage-knowledge} for use in robots.
  Considering the fluid exudation function of cartilage in living organisms, the requirements for mimicking cartilage are summarized as follows:
  \begin{itemize}
    \item Exudation of synovial fluid when subjected to load.
    \item Gradual increase in synovial fluid exudation with increasing load (not exuding all synovial fluid at once).
    \item Reusable in repetitive applications.
  \end{itemize}
  In addition to these conditions, considering the application to biomimetic robots, it is desirable for the mechanism to fulfill the following:
  \begin{itemize}
    \item Cost-effectiveness.
    \item Capability for 3D printing to enable free design.
  \end{itemize}

  We propose the mechanism depicted in \figref{fig:cartilage-overview} as a model for fluid-exuding cartilage that fulfills these requirements.
  The cartilage section can be 3D printed at a low cost using rubber material, and absorbent materials are vertically inserted at regular intervals on the load-bearing surface.
  Furthermore, an absorbent material layer is provided at the base.
  The joint itself is covered by the joint capsule, filled with synovial fluid, so that synovial fluid is present on the cartilage surface under steady-state conditions.
  
  STEP1: When a load is applied, the synovial fluid on the cartilage surface becomes a thin layer.
  
  STEP2: In the initial stage of load application, the absorbent material layer of the soft base compresses.
  
  STEP3: After the compression of the absorbent material, the rubber part, which is the cartilage part, further compresses and deforms according to the load.
  The rubber material compresses in the direction of the load, simultaneously deforming to press against the insertion of the absorbent material from the side.
  As rubber compresses according to the load, with a Poisson's ratio of approximately 0.5, it deforms the absorbent material insert from the side by about half the volume of the compressive amount in the direction of the load.
  As a result, synovial fluid exudes from the absorbent material insert, combining the amount compressed in the direction of the load and the amount deformed by being pushed against the rubber from the side.
  The exuded synovial fluid increases with the load.
  The base has already been compressed and is in a state where synovial fluid is difficult to pass through, so the exuded synovial fluid is also pushed out to the load surface and serves as a lubricating function.
  
  STEP4: When the load is removed, the rubber restores its shape elastically, and at the same time, the absorbent material layer at the base absorbs synovial fluid from the surroundings, replenishing the synovial fluid to the absorbent material insert of the cartilage part.
  The reabsorption of synovial fluid during load removal enables reuse.

  Regarding the lubrication theory of cartilage, various studies have been conducted, but a definitive theory has not yet been concluded.
  In \figref{fig:cartilage-overview}, for clarity, diagrams similar to the concepts of weeping lubrication\cite{mccutchen1962weeping-lubrication} and biphasic lubrication theory\cite{mow1980biphasic} are presented.
  Weeping lubrication was once questioned for the idea that synovial fluid would move to high-pressure friction surfaces, and there was a period when it was not supported.
  However, it has come to be supported in a new form called biphasic lubrication theory, which posits that both the solid phase and the liquid phase exert influence in response to external loads.
  In biphasic lubrication theory, it is argued that due to the low permeability of cartilage, there is no place for the movement of fluid, and the fact that unlike rubber, which undergoes elastic deformation and compression, fluid is incompressible and therefore maintains a constant volume, leads to the accumulation of trapped fluid, which is eventually pushed out to the friction surface to bear the load support.
  It is also argued that as the load support from the fluid increases, the coefficient of friction decreases.
  A detailed review of biphasic lubrication theory is presented in \cite{ateshian2009biphasic-review}.
  While biphasic lubrication theory is not definitive, the liquid exudation mechanism proposed in this study may replicate the lubrication function occurring at a microscopic scale in 3D printing at a macroscopic scale.

  The choice of absorbent material for the cartilage section and the spacing for inserting absorbent material into the cartilage part are verified in \secref{sec:pre-experiment}.
  In the following sections, we conduct experiments using the created fluid-exuding cartilage sheet to examine the above items and to verify its functionality and usefulness.

\begin{figure}[t]
  \begin{center}
    \includegraphics[width=\linewidth]{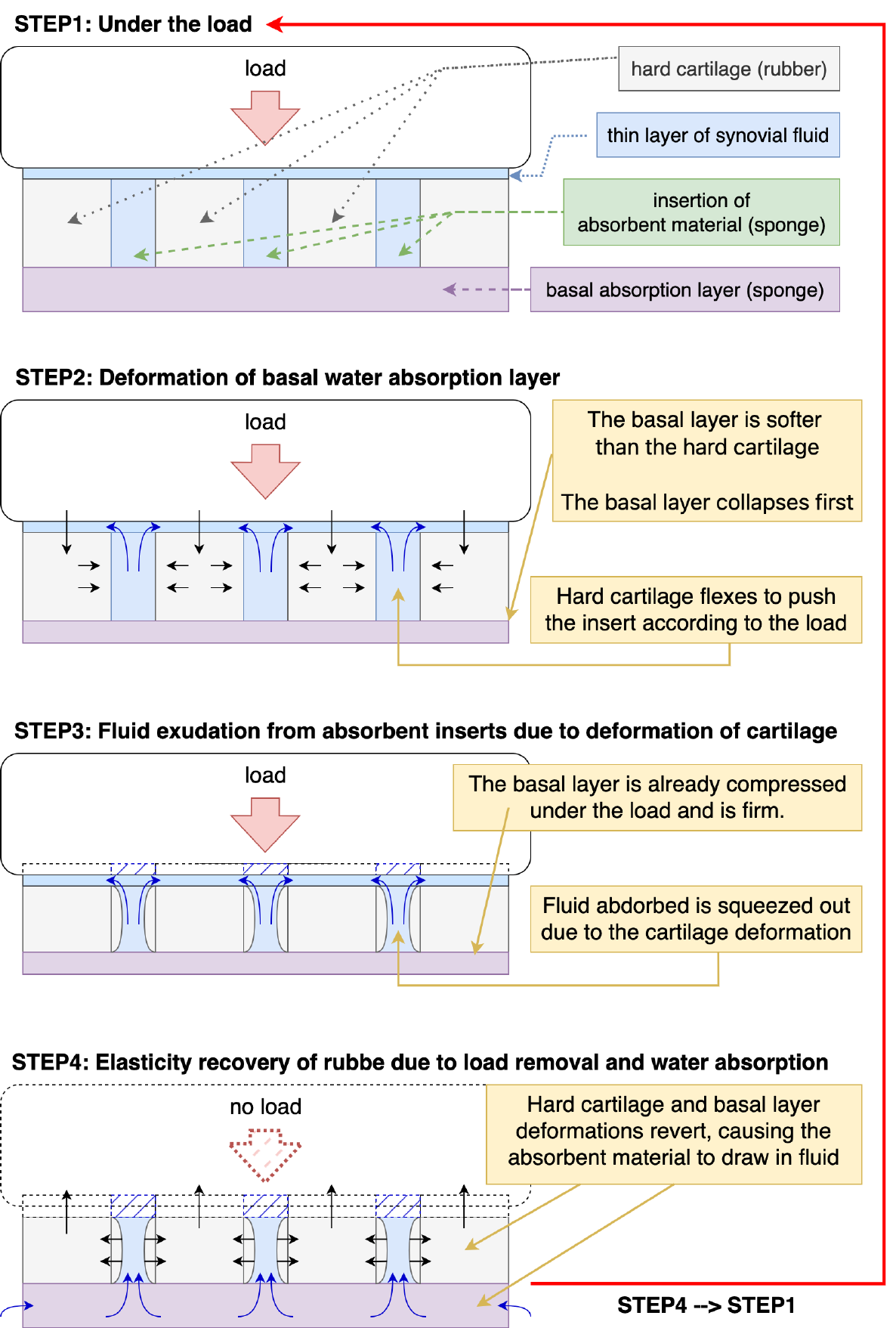}
    \caption{Figure of the fluid-exuding cartilage mechanism.}
    \label{fig:cartilage-overview}
  \end{center}
\end{figure}
}%
{%
第3章で述べた生体軟骨の特性をロボットにおいて模倣再現する機構を本章では提案する．
生体での軟骨の液体滲出機能を考慮すると，模倣再現する軟骨の要件は以下のようにまとめられる．
\begin{itemize}
  \item 荷重がかかると関節液が滲出されること
  \item 荷重負荷量が増えるにつれて関節液の滲出量も増えること（一度に関節液を全て滲出しない）
  \item 反復利用が可能であること
\end{itemize}
これらの条件に加えて，生体模倣ロボットへの適用を考慮すると，さらに以下の条件も満たすことが望ましい．
\begin{itemize}
  \item 低コストであること
  \item 自由な設計を可能とするために3Dプリンティングできること
\end{itemize}

これらの要件を満たす液体滲出軟骨の機構として，\figref{cartilage-overview}のようなメカニズムを提案する．
軟骨部はゴム素材を用いることで低コストに3Dプリンティングでき，吸水性素材が荷重負荷面に対して垂直に一定間隔で挿入される．
さらに，基底部にも吸水性素材の層が設けられる．
関節全体は関節包に覆われ，中に関節液が満ちるため，定常時には軟骨表面に関節液が存在する．

液体滲出機構の機序は以下である．

\textcolor{red}{
  STEP1:}
荷重がかかり始めると，軟骨表面の関節液は薄い層となる．
\textcolor{red}{
  STEP2:}
荷重負荷の初期段階では，柔らかい基底部の吸水性素材の層が圧縮される．
\textcolor{red}{
  STEP3:}
吸水性素材の圧縮後には，さらに軟骨部であるゴム素材が荷重に応じて圧縮変形していく．
ゴム素材は，荷重方向に対して圧縮されると同時に，吸水性素材の挿入部を側面から押すようにも変形する．
ゴムは荷重量に応じて圧縮変形し，ゴムのポアソン比は約0.5であるため，荷重方向の圧縮量の約半分の体積だけ吸水性素材挿入部を横から押すように変形する．
その結果，吸水性素材挿入部からは，ゴム変形とともに荷重方向に対して圧縮される量と，荷重に応じて側面からゴムに押されて変形する量を合わせた変形量の関節液が滲出される．
液体滲出量は荷重に応じて増えていく．
\textcolor{red}{
 基底部はすでに圧縮されて関節液が通過しにくい状態となっているため，滲出された関節液は荷重面へも押し出され，潤滑機能を担う．
}
\textcolor{red}{
  STEP4:}
荷重負荷が除去されると，ゴムが弾性変形から復元すると同時に，基底部の吸水性素材層が周囲から関節液を吸収し，軟骨部の吸水性素材挿入部にも関節液が再供給される．
荷重負荷除去時に関節液を再吸収することから，反復利用も可能となる．

\textcolor{red}{
 軟骨の潤滑理論に関しては，様々な研究が行われてものの，完全な理論はまだ結論づけられていない．
 \figref{cartilage-overview}では，わかりやすさのため，滲出潤滑\cite{mccutchen1962weeping-lubrication}や二相性潤滑説\cite{mow1980biphasic}の考え方に近い図とした．
 滲出潤滑は，流体である関節液がわざわざ高圧の摩擦面に移動することについて疑問視され，支持されない時期もあった．
 しかし，外荷重に対して固相と液相の両者が効力を発揮しているという二相性潤滑説という新たな形で支持されるようになってきている．
 二相性潤滑説では，軟骨の透水率の低さによって液体の移動場所がないことと，弾性変形して圧縮されるゴムと違って液体は非圧縮性をもつために体積が一定であることの２つの要因によって，逃げ場を失った流体が蓄積して結果的に摩擦面に押し出されて荷重支持を担うとされる．
 流体の荷重支持分が大きくなるにつれて摩擦係数も減少するとされる．
 二相性潤滑説の詳しいレビューは\cite{ateshian2009biphasic-review}にて近年紹介されている．
 二相性潤滑説が完全に正しいとは断定できないが，いずれにせよ，本研究で提案する液体滲出機構は，ミクロなスケールで行われている液体滲出機能を3Dプリンティングで再現できるマクロなスケールで実現する機構と解釈できる．
}

軟骨部に対してどのような吸水性素材を用いるかについては，
第5章
にて検証する．
また，どのような間隔で軟骨部に吸水性素材を挿入するかについても
第5章
にて検証する．
次章以降では，実際に作成した液体滲出軟骨シートを用いて，実験を行い，
上記の項目の検討に加えて，機能や有用性についても検証する．

\begin{figure}[t]
  \begin{center}
    \includegraphics[width=\linewidth]{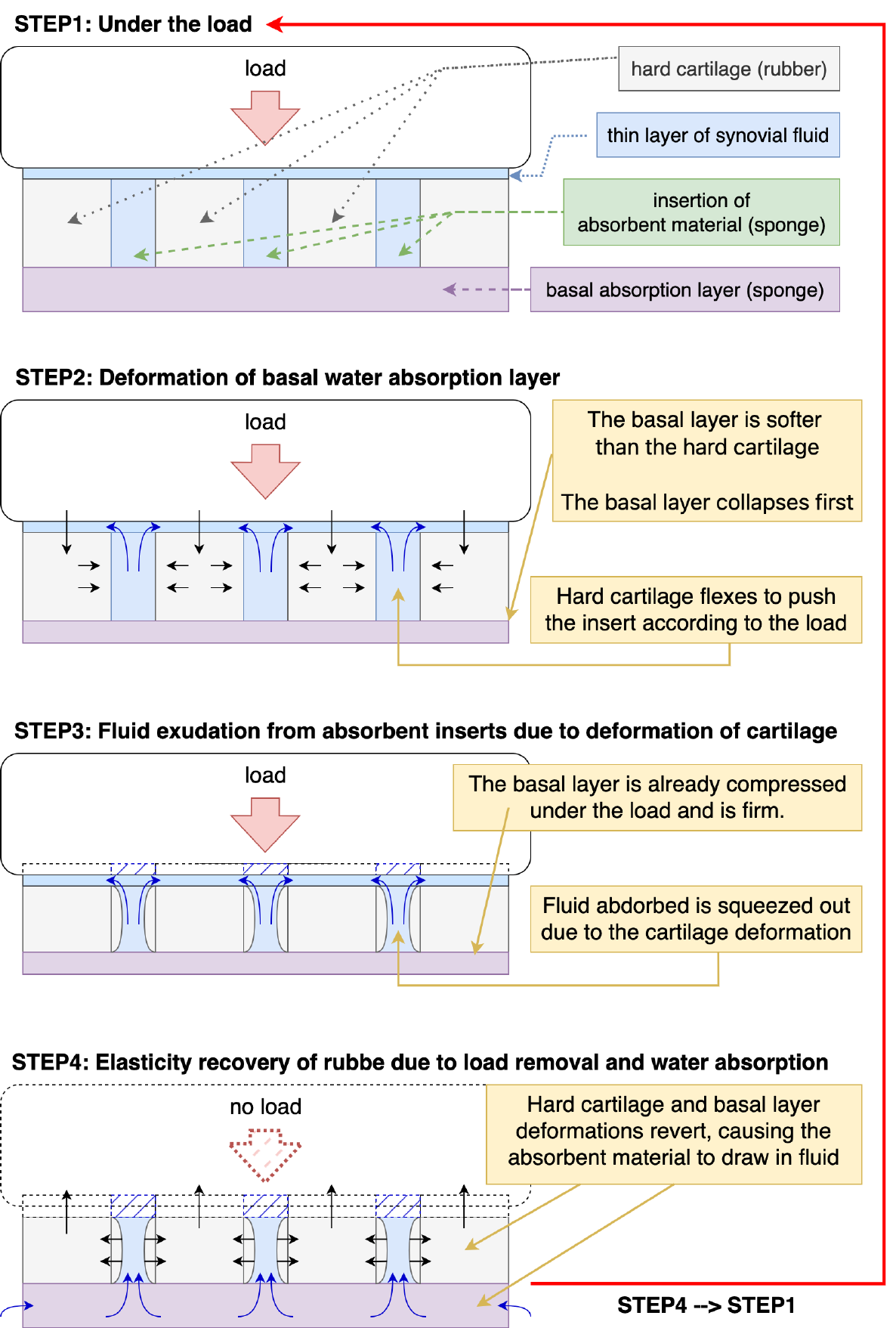}
    \caption{Figure of the fluid-exuding cartilage mechanism.}
    \label{fig:cartilage-overview}
  \end{center}
\end{figure}
}%

\section{Preliminary Experiment: Fluid Exudation Experiment under Load Conditions}
\label{sec:pre-experiment}
\switchlanguage%
{%
  As a preliminary experiment, we inserted \SI{0.5}{mm}-diameter kite strings, \SI{1.6}{mm}-diameter kite strings, and PVA sponges with a size of 2-\SI{3}{mm} square at regular intervals perpendicular to the load-bearing surface into a rubber molding sheet.
  We then verified whether fluid would exude when a load was applied.
  Kite strings were chosen for their capillary properties, and PVA sponges were selected for their absorbent and flexible nature.
  Holes corresponding to each string's diameter were pre-made in the rubber during 3D printing, and the strings were sewn using a needle.
  Similarly, holes of the appropriate size for the inserted sponge were made during rubber 3D printing, and cut pieces of sponge were inserted.
  The rubber sheet was created to be approximately \SI{5}{mm} thick, similar to the thickness of human cartilage.
  The rubber material used was Elastic50A from Formlabs, which is the softest material they offer and is readily 3D printable.
  Elastic50A has a Shore hardness of 50A, and according to the relationship between Shore hardness and Young's modulus from previous studies \cite{Gent1958OnTR}, the Young's modulus is around $\SI{3.0}{MPa}$.
  Although slightly larger than the Young's modulus of cartilage, it is relatively close, and the advantage of being able to 3D print at low cost led to its adoption.

  The procedure for the preliminary experiment is as follows.
  First, hyaluronic acid solution was diluted to approximately 3-\SI{4}{mg/ml}, similar to the viscosity of human synovial fluid, to create a pseudo-synovial fluid.
  Additionally, the created synovial fluid was dyed red with food coloring to visualize fluid exudation.
  Strings and sponges were inserted into the created fluid-exuding cartilage sheet, which was immersed in colored synovial fluid, to ensure sufficient impregnation of synovial fluid into the internal strings and sponges.
  After wiping off the surface moisture of the cartilage sheet, a paper wiper was placed on the cartilage sheet as shown in \figref{fig:pre-experiment-overview}, and a load was applied on top using an 8-pound dumbbell ($\approx$\SI{3.6}{kg}).
  The degree of red coloration on the paper wiper after applying the load was used to verify synovial fluid exudation.

\begin{figure}[t]
  \begin{center}
    \centering
    \includegraphics[width=0.8\columnwidth]{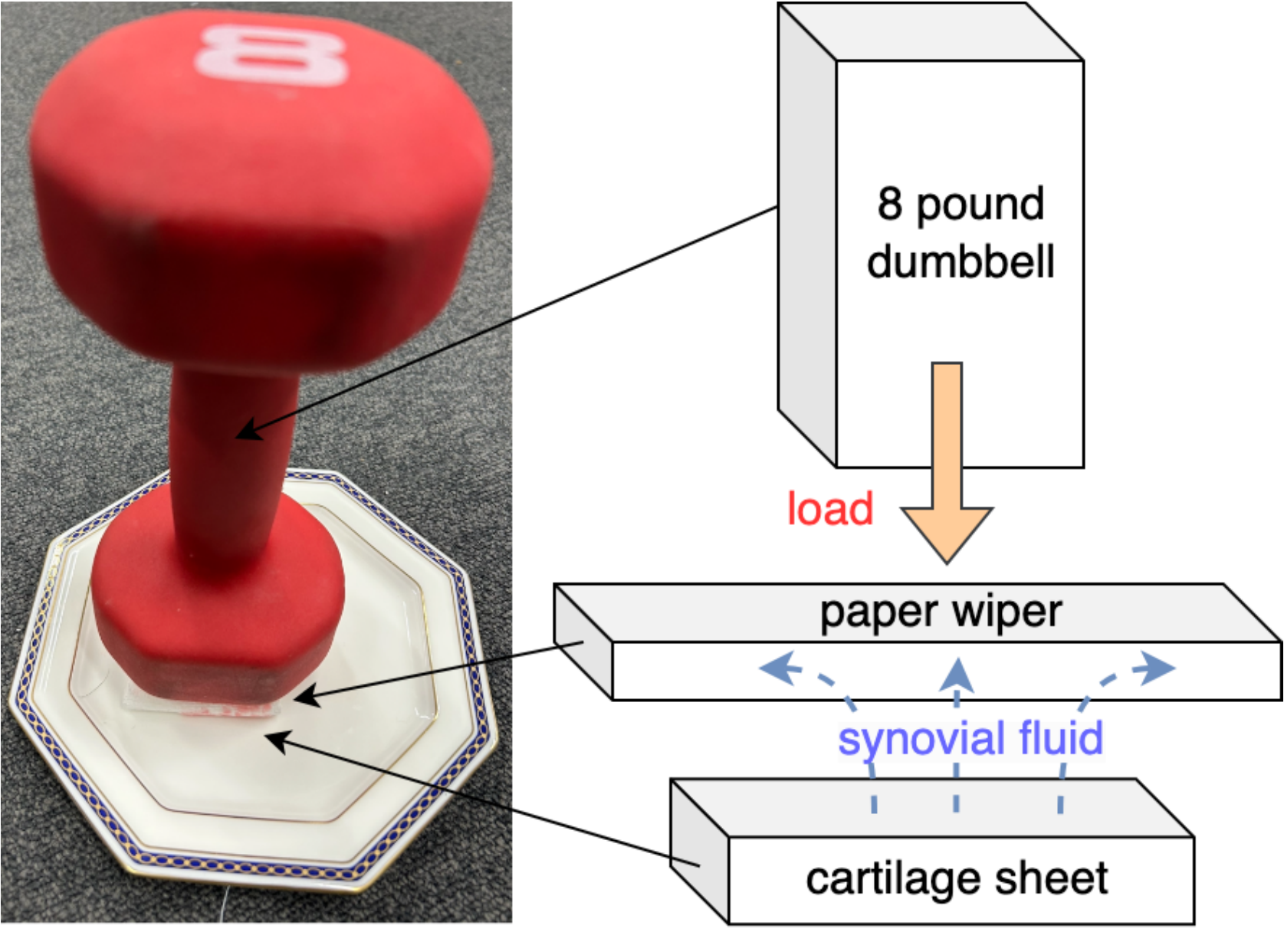}
    \caption{Overview of fluid exudation experiment under load conditions.}
    \label{fig:pre-experiment-overview}
  \end{center}
\end{figure}

  The results are shown in \figref{fig:pre-experiment-0.5mm}, \figref{fig:pre-experiment-1.6mm}, \figref{fig:pre-experiment-pva-8pound}.
  With kite strings, regardless of diameter, sufficient exudation was not observed.
  The lack of sufficient moisture retention in the strings and the hardness of the strings surpassing the deforming force of the rubber are considered possible reasons.
  In contrast, fluid exudation was confirmed with sponges.

\begin{figure}[t]
  \begin{tabular}{cc}
    \begin{minipage}[b]{.33\columnwidth}
      \centering
      \captionsetup{width=0.9\columnwidth}
      \includegraphics[width=0.75\columnwidth]{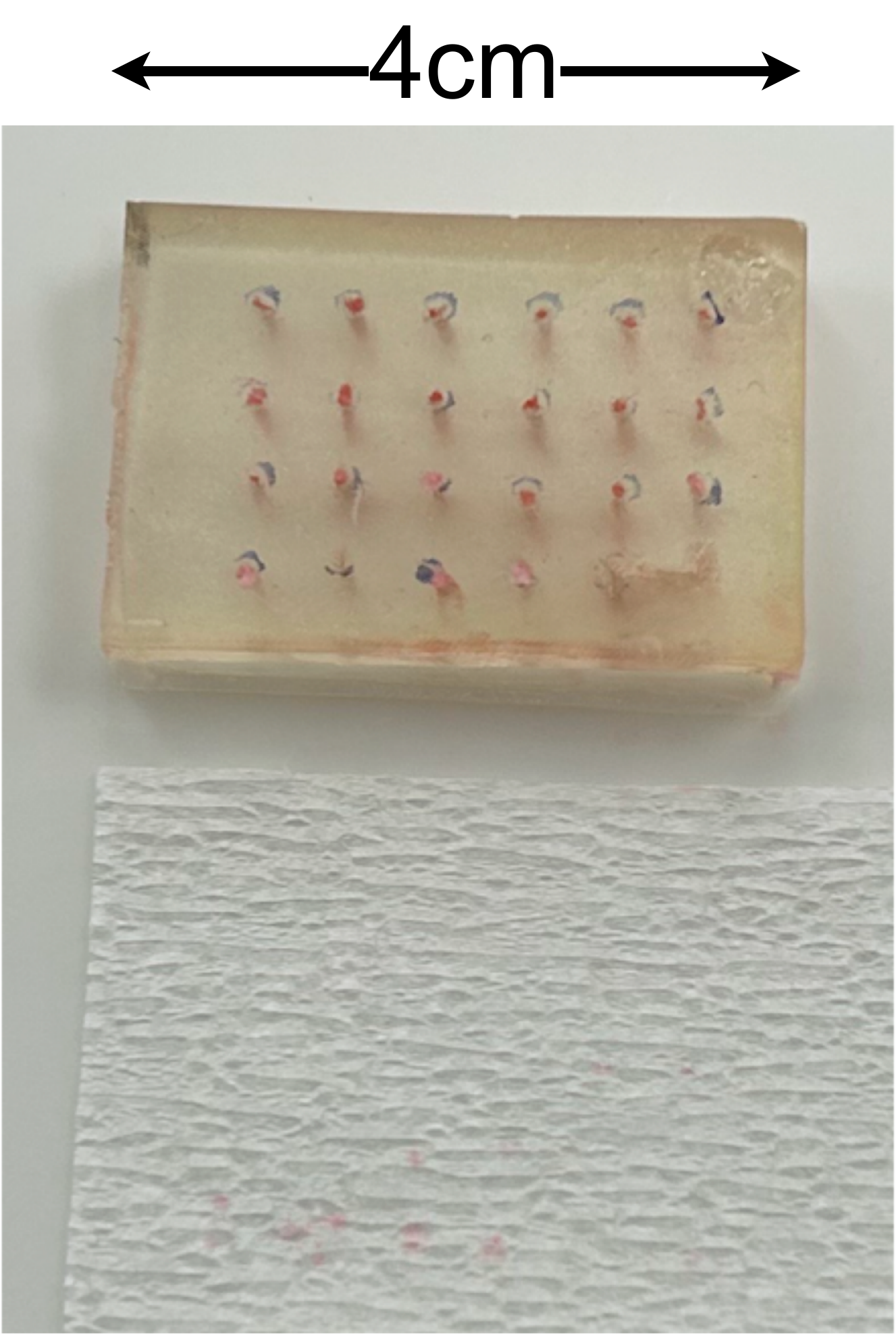}
      \caption{Fluid exudation experiment with \SI{0.5}{mm} diameter kite string (8-pound load).}
      \label{fig:pre-experiment-0.5mm}
    \end{minipage}
    \begin{minipage}[b]{.33\columnwidth}
      \centering
      \captionsetup{width=0.9\columnwidth, justification=raggedright}
      \includegraphics[width=0.75\columnwidth]{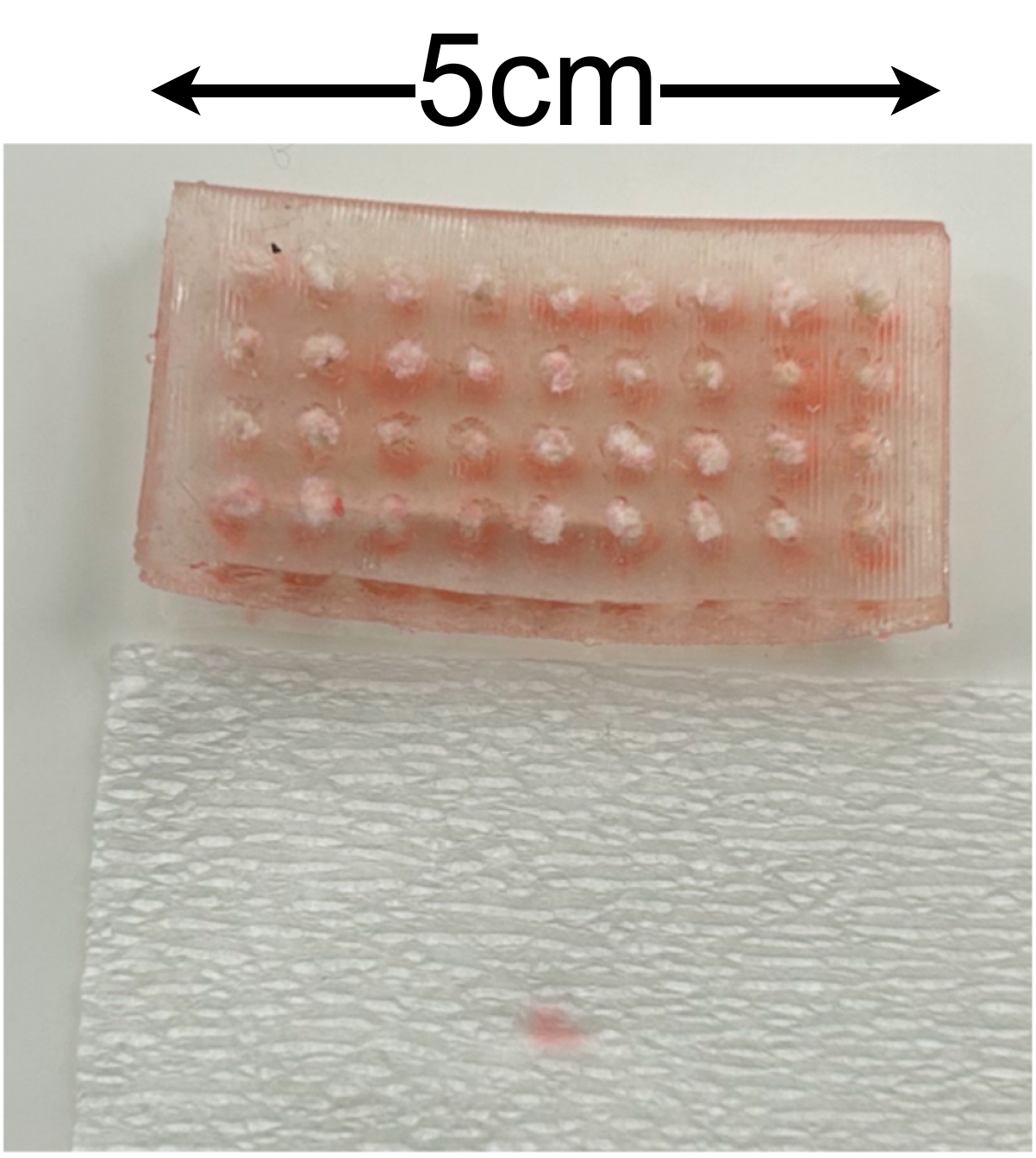}
      \caption{Fluid exudation experiment with \SI{1.6}{mm} diameter kite string (8-pound load).}
      \label{fig:pre-experiment-1.6mm}
    \end{minipage}
    \begin{minipage}[b]{.33\columnwidth}
      \captionsetup{width=0.9\columnwidth, justification=raggedright}
      \centering
      \includegraphics[width=0.75\columnwidth]{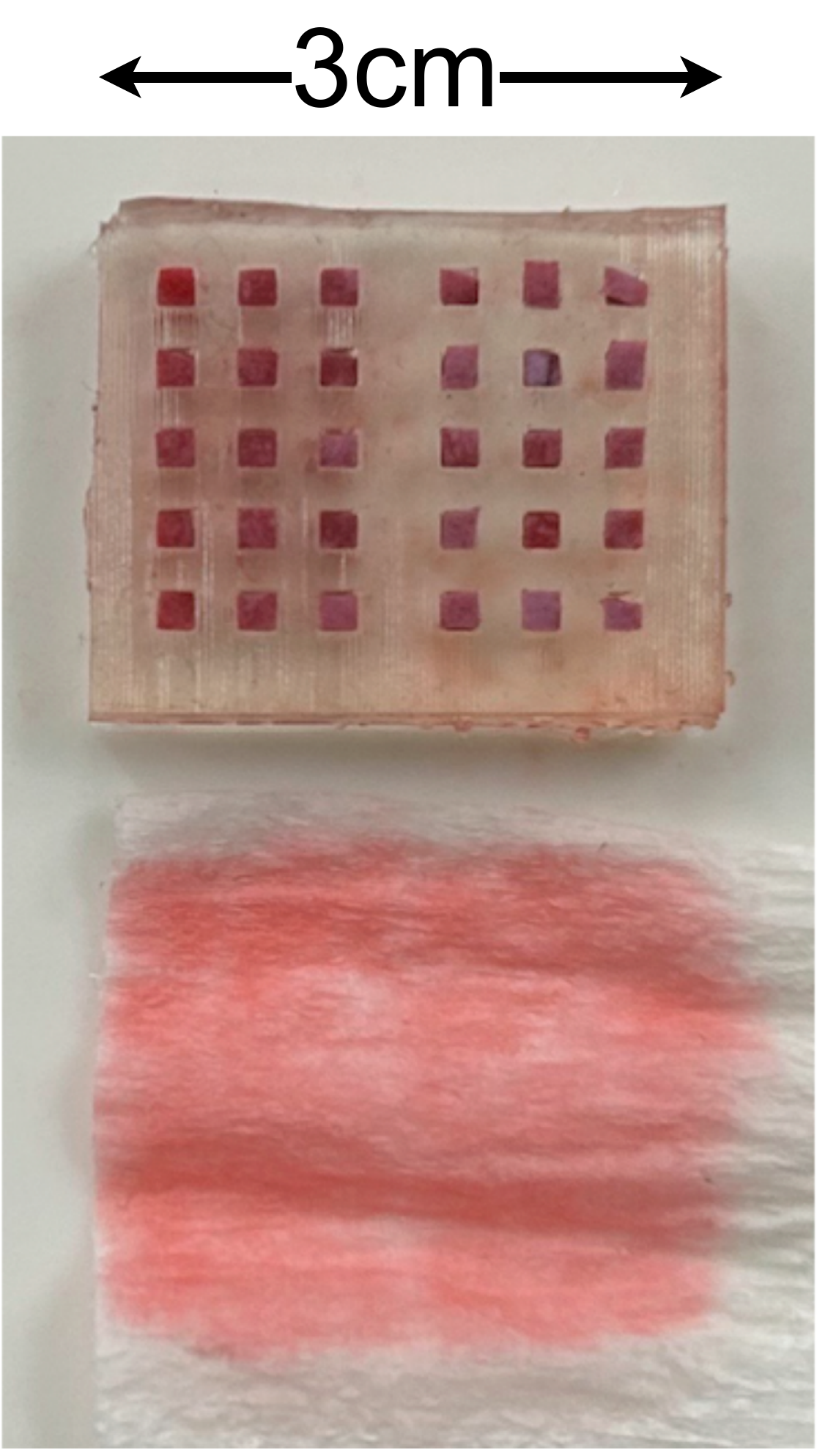}
      \caption{Fluid exudation experiment with PVA sponge (8-pound load).}
      \label{fig:pre-experiment-pva-8pound}
    \end{minipage}
  \end{tabular}
\end{figure}

  For the cartilage sheet created using sponges, additional experiments were conducted to observe changes in exudation with varying load.
  After applying an 8-pound dumbbell ($\approx$\SI{3.6}{kg}) load and repeating the operation of absorbing exudation with a paper wiper several times, the amount absorbed by the wiper decreased, as depicted in \figref{fig:pre-experiment-pva-8pound-little-exudation}.
  From the state shown in \figref{fig:pre-experiment-pva-8pound-little-exudation}, the load was increased to 15 pounds ($\approx$\SI{6.8}{kg}).
  As the deformation of the rubber increased, it was anticipated that the amount of fluid exudation would also increase.
  The results, shown in \figref{fig:pre-experiment-pva-15pound-more-exudation}, confirmed further fluid exudation corresponding to the applied load.

\begin{figure}[t]
  \begin{tabular}{cc}
    \begin{minipage}[t]{.48\columnwidth}
      \centering
      \captionsetup{width=.9\columnwidth}
      \includegraphics[width=0.75\columnwidth]{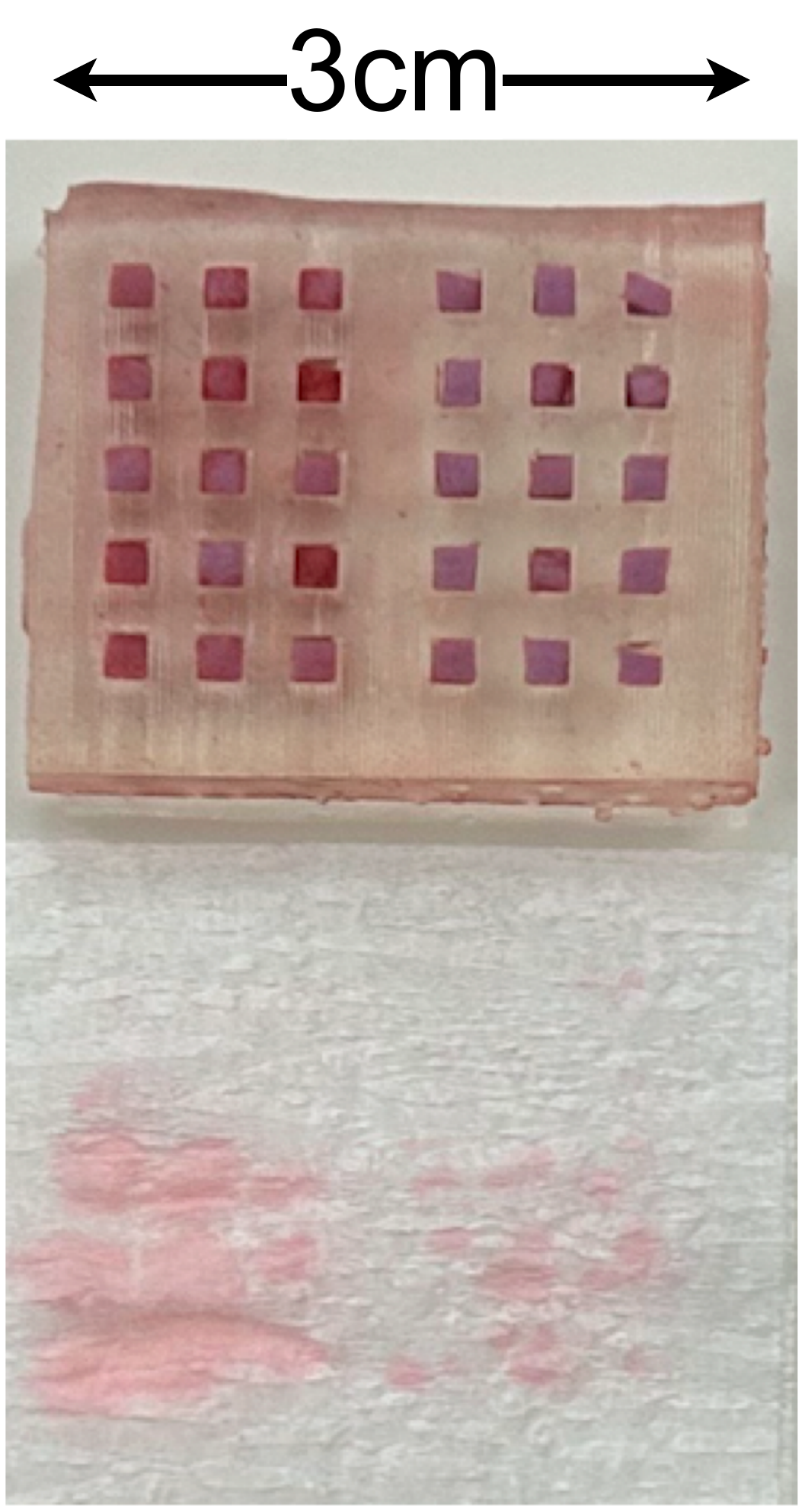}
      \caption{Fluid exudation experiment with PVA sponge after many times (8-pound load).}
      \label{fig:pre-experiment-pva-8pound-little-exudation}
    \end{minipage}
    \begin{minipage}[t]{.48\columnwidth}
      \centering
      \captionsetup{width=.9\columnwidth}
      \includegraphics[width=0.75\columnwidth]{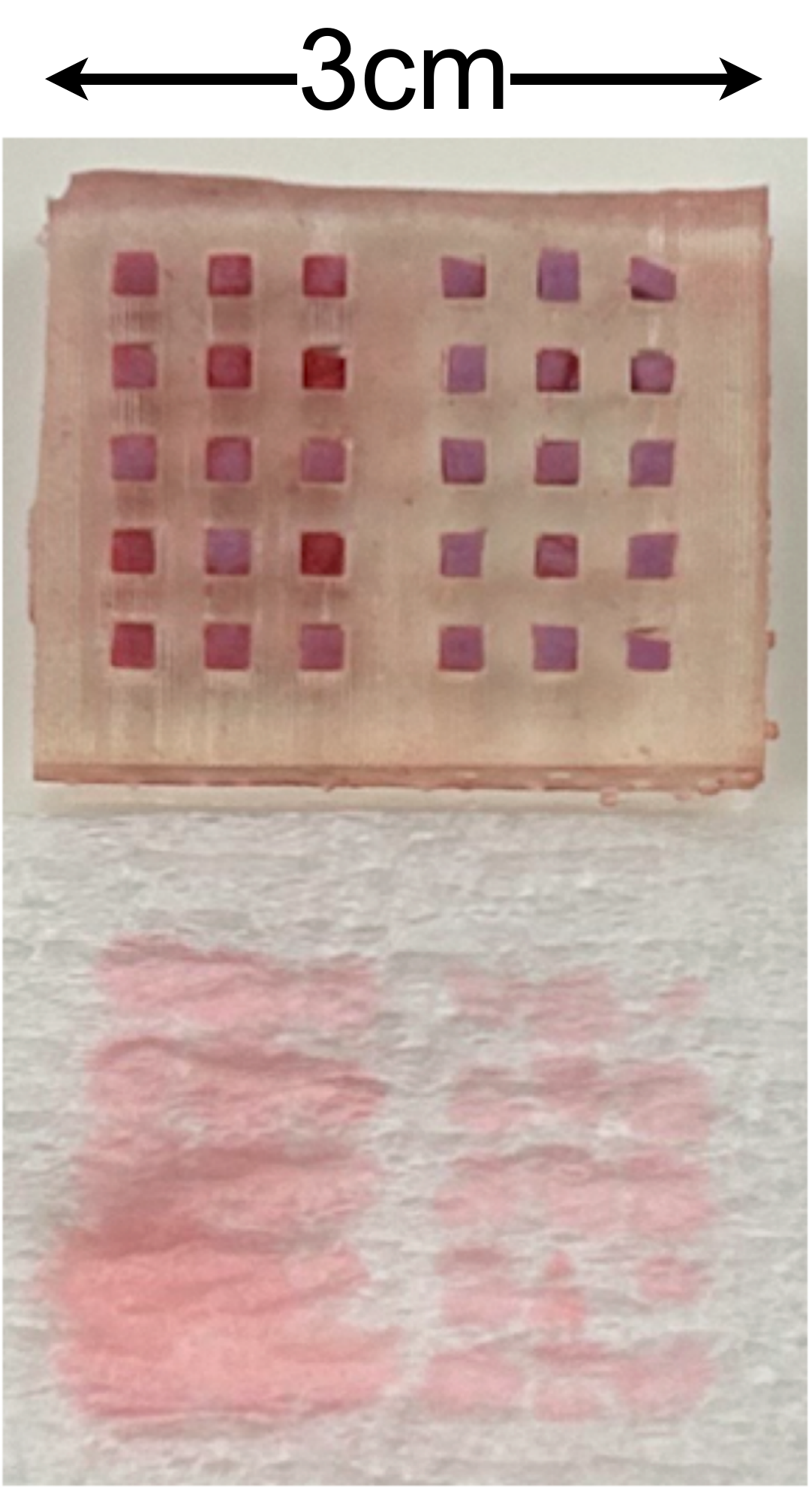}
      \caption{Fluid exudation experiment with PVA sponge (15-pound load after \figref{fig:pre-experiment-pva-8pound-little-exudation}).}
      \label{fig:pre-experiment-pva-15pound-more-exudation}
    \end{minipage}
  \end{tabular}
\end{figure}

  Furthermore, we compared the retention of fluid exudation performance when repeatedly applying a load between cases where a sponge was laid out like a sheet at the bottom of the cartilage sheet and cases where it was not.
  After applying an 8-pound dumbbell ($\approx$\SI{3.6}{kg}) load, the operation of absorbing with a paper wiper was repeated five times, and the amount of fluid exudation onto the paper wiper after the fifth operation was compared.
  Between each operation, the state of no load was interleaved, and the paper wiper was replaced each time.
  The results are shown in \figref{fig:pre-experiment-pva-8pound-repeat-without-sheet} and \figref{fig:pre-experiment-pva-8pound-repeat-with-sheet}.
  Fluid exudation functionality was retained when laying out a sponge sheet at the bottom of the cartilage sheet.
  It was observed that the sponge sheet at the bottom of the cartilage sheet absorbed synovial fluid when there was no load, supplying synovial fluid to the cartilage sheet.

\begin{figure}[t]
  \begin{tabular}{cc}
    \begin{minipage}[b]{.48\columnwidth}
      \centering
      \captionsetup{width=.9\columnwidth}
      \includegraphics[width=0.75\columnwidth]{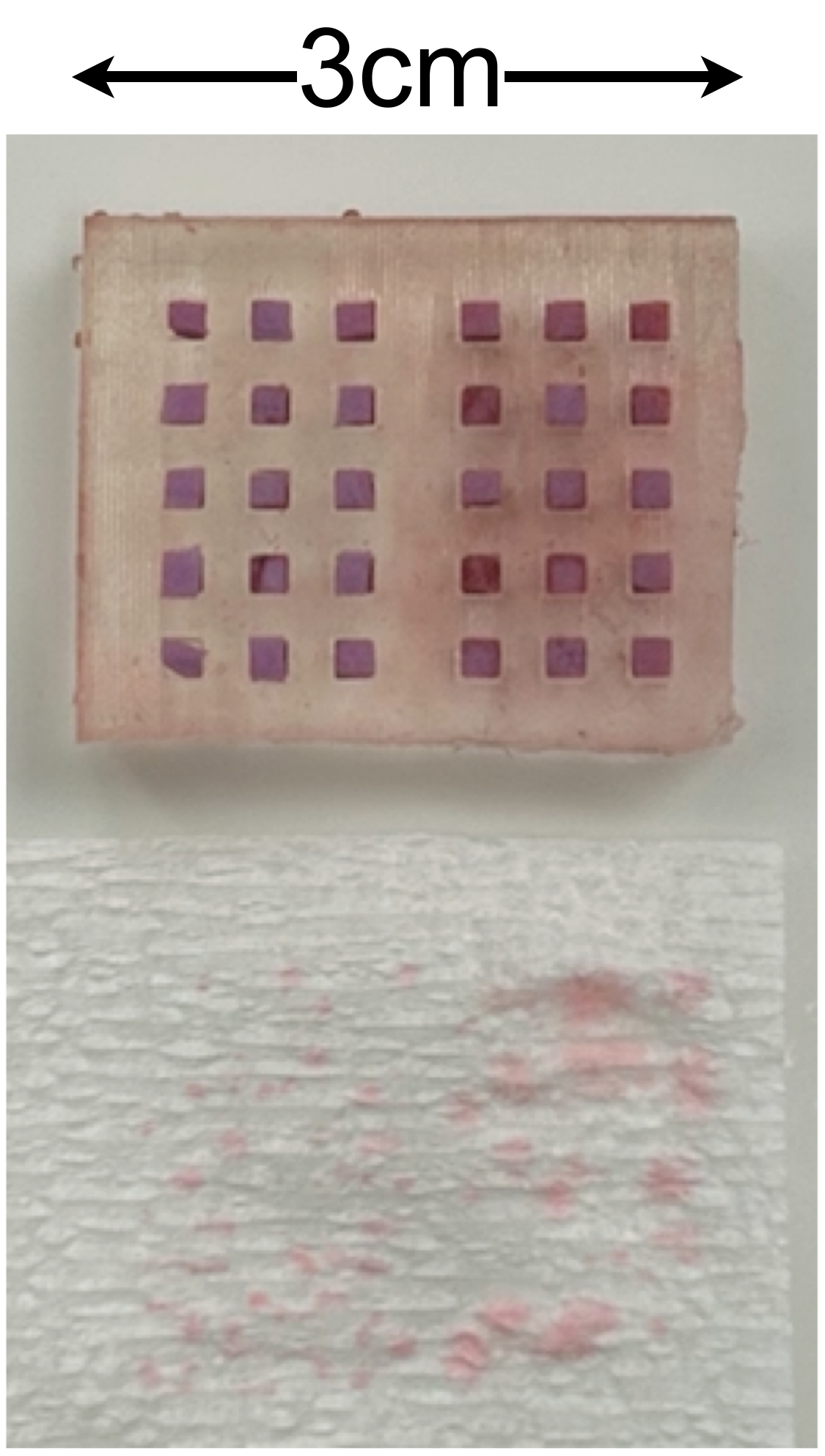}
      \caption{Fluid exudation experiment using PVA sponge after applying 8-pound load 5 times without base sheet.}
      \label{fig:pre-experiment-pva-8pound-repeat-without-sheet}
    \end{minipage}
    \begin{minipage}[b]{.48\columnwidth}
      \centering
      \captionsetup{width=.9\columnwidth}
      \includegraphics[width=0.75\columnwidth]{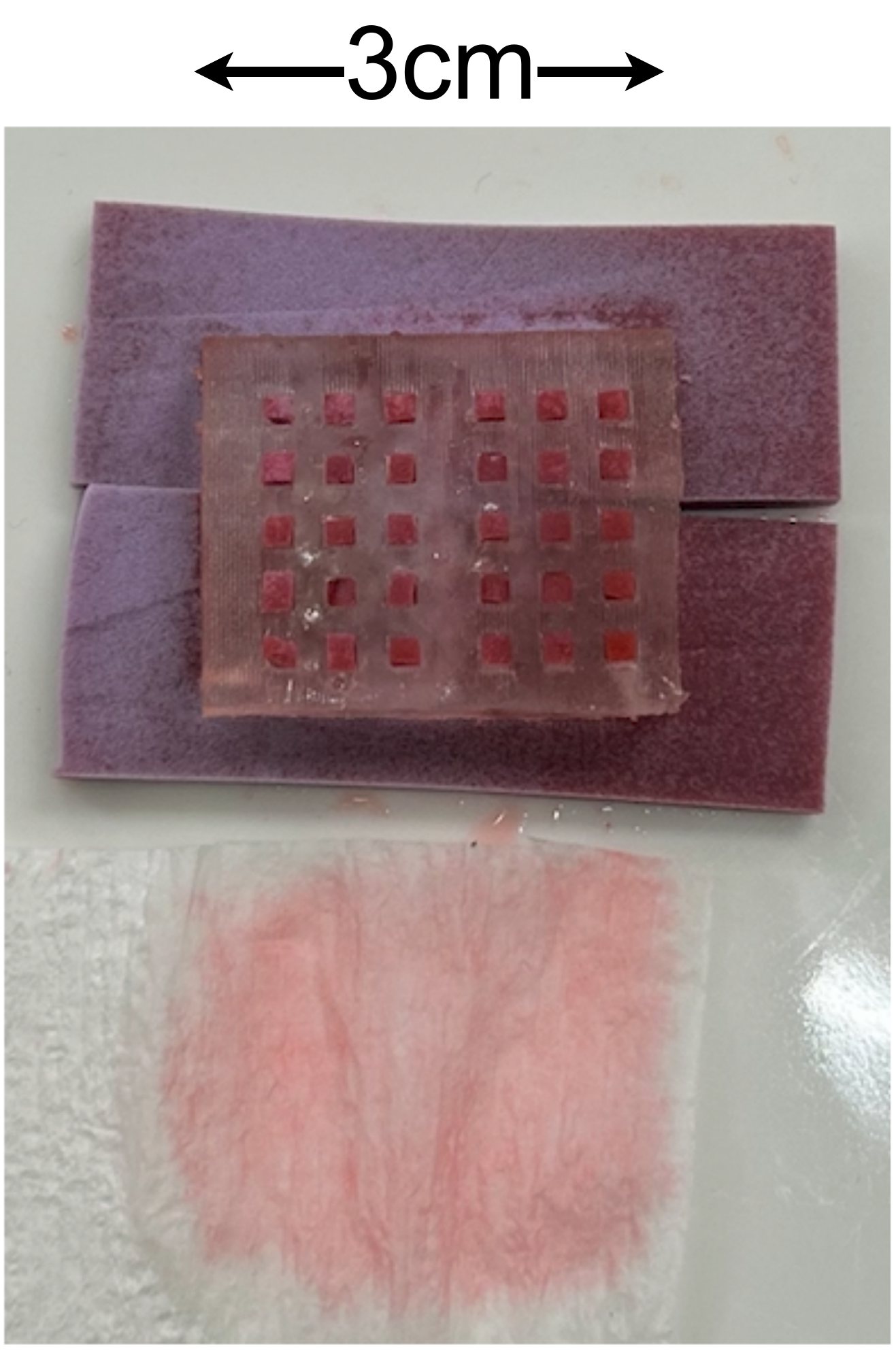}
      \caption{Fluid exudation experiment using PVA sponge after applying 8-pound load 5 times with base sheet.}
      \label{fig:pre-experiment-pva-8pound-repeat-with-sheet}
    \end{minipage}
  \end{tabular}
\end{figure}

  From these results, it was confirmed that the cartilage sheet created using a sponge has the function of exuding fluid under load.
  Additionally, it was observed that the amount of fluid exudation increases with the load.
  Furthermore, it was confirmed that the fluid exudation functionality is retained by laying out a sponge sheet at the bottom of the cartilage.

  Considering the microstructure inside living cartilage, it is desirable to insert small-sized materials in large quantities.
  As a size that can be handcrafted, we decided to insert absorbent materials of \SI{2}{mm} square.

  By considering the curvature of the cartilage, it is possible to design the cartilage sheet to always have the fluid-absorbent material insertion points in contact with the load surface when the cartilage sheet undergoes compressive deformation due to load loading.
  Calculations using trigonometric functions indicate that placing the fluid-absorbent material insertion points at \SI{2}{mm} intervals is sufficient.
  Further details are explained below.

  Assuming the curvature of the cartilage is denoted as $1/r$, the cartilage surface forms a virtual circle with a radius $r$.
  Under the load loading, it is assumed that the cartilage undergoes elastic deformation of $\delta$ mm, as shown in \figref{fig:cartilage-surface-under-load}.
\begin{figure}[t]
  \begin{center}
    \centering
    \includegraphics[width=0.7\columnwidth]{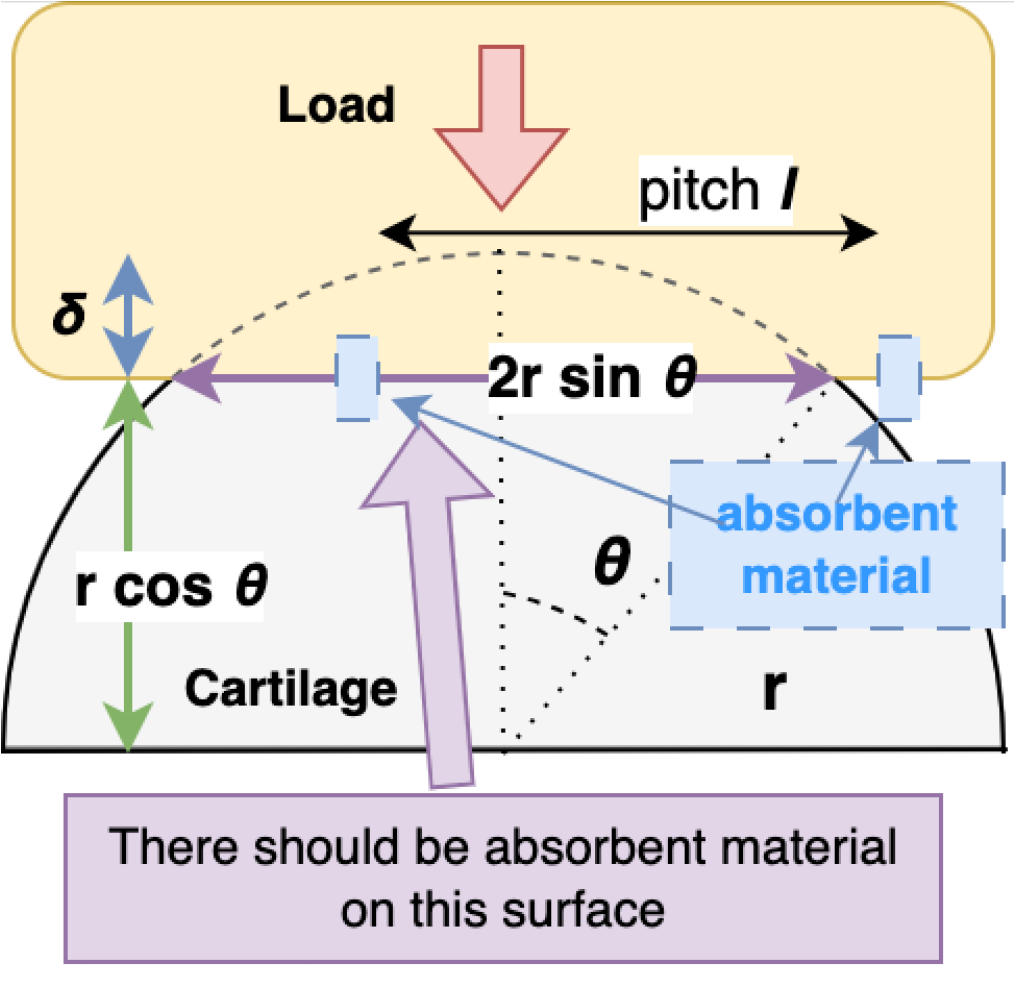}
    \caption{Figure of cartilage surface under load conditions.}
    \label{fig:cartilage-surface-under-load}
  \end{center}
\end{figure}
  In this case, when the pitch interval of the fluid-absorbent material insertion points is denoted as $l$ mm, the following equation needs to hold for the fluid-absorbent material insertion points to always exist on the plane formed during elastic deformation.
\begin{align}
  r\cos(\theta) &= r - \delta \\
  2r\sin(\theta) &\ge l \label{eq:pitch-condition}
\end{align}
Here, $\theta$ represents the angle parameter, indicating the rotation angle on the virtual circle formed by the cartilage surface.

The relationship expressed by \equref{eq:pitch-condition} can be rearranged as follows:
\begin{align}
  l &\le 2r\sin(\theta) \notag \\
    &\le 2r\sqrt{1 - (\cos(\theta))^2} \notag \\
    &\le 2\sqrt{r^2 - (r-\delta)^2} \label{eq:pitch}
\end{align}

In practical terms, the deflection amount $\delta$ of the cartilage sheet under high-load conditions, where fluid-exuding cartilage is expected to demonstrate its effectiveness, is approximately \SI{1}{mm}.
In an actual cartilage sheet with \SI{2}{mm} square fluid-absorbent material insertion points placed at approximately \SI{2}{mm} intervals, it underwent approximately \SI{1}{mm} compression deformation when a weight of a 15-pound dumbbell ($\approx\SI{6.8}{kg}$) was applied.
This can be considered a sufficiently plausible load even when considering the daily movements of the human body.

Considering the curvature of the cartilage, given the size of human bones, it can be approximated that $r\geq\SI{3}{mm}$.
In this case, the square root term in \equref{eq:pitch} is greater than or equal to 1.
Therefore, under the load conditions, if $l<=\SI{2}{mm}$, the condition for the fluid-absorbent material insertion points to always exist on the plane formed during elastic deformation is satisfied.

Based on the above discussions, in subsequent experiments, a cartilage sheet with \SI{2}{mm} square fluid-absorbent material insertion points placed at approximately \SI{2}{mm} intervals will be utilized.

}%
{%
予備実験として，ゴム造形シートの中に太さ\SI{0.5}{mm}の凧糸，太さ\SI{1.6}{mm}の凧糸，2$\sim$\SI{3}{mm}平方のPVAスポンジを，荷重面に対して一定間隔で垂直に挿入し，荷重をかけた際に液体が滲出するかを検証した．
凧糸はそれ自身が持つ毛細管としての性質を利用することを意図して採用し，PVAスポンジは吸水性と柔軟性を利用することを意図して採用した．
凧糸に関しては，それぞれの太さに相当する穴をゴムの造形時に開けておき，針で縫い付けた．
スポンジに関しても，挿入するスポンジに相当する大きさの穴をゴムの造形時に開けておき，小さくカットしたスポンジを挿入した．
ゴム造形シートは人間の軟骨の厚さに近い\SI{5}{mm}程度で作成した．
ゴム素材にはFormlabs社で最も柔らかい素材であるElastic50Aを用いた．
Elastic50Aは手軽に3Dプリンティング可能なレジンゴムの中で比較的柔らかい素材であり，ショア硬度が50Aである．
先行研究のショア硬度とヤング率の関係\cite{Gent1958OnTR}によるとヤング率は約\SI{3}{MPa}付近となる．
軟骨のヤング率よりも多少大きな値ではあるが，比較的近い値であり，また低コストで3Dプリンティングできる利点を持つため，採用した．

予備実験の手順は以下である．
まず，ヒアルロン酸水溶液を人体の関節液と同程度の3$\sim$\SI{4}{mg/ml}に薄めて，疑似関節液を作成する．
また，作成した関節液に食紅を溶かして赤に染色し，液体の滲出を可視化できるようにしておく．
さらに，糸やスポンジを挿入して作成した液体滲出軟骨シートを，色のついた関節液に浸し，内部の糸やスポンジに十分に関節液を染み込ませる．
そして，軟骨シートの表面の水分を拭き取った上で，\figref{fig:pre-experiment-overview}のように紙ワイパーを軟骨シートの上にのせ，その上から荷重をかける．
荷重には8ポンドのダンベルを利用した．
荷重をかけた後の紙ワイパーへの赤色の付着度合いで，関節液滲出を検証した．

結果は\figref{fig:pre-experiment-0.5mm}，\figref{fig:pre-experiment-1.6mm}，\figref{fig:pre-experiment-pva-8pound}のようになった．
凧糸では太さによらず，十分な量の滲出を確認することができなかった．
凧糸は，水分保有量が十分でないことや，ゴムが変形しようとする力よりも糸の硬さが上回っていることなどが原因として考えられる．
対して，スポンジでは液体の滲出を確認することができた．

\begin{figure}[t]
  \begin{center}
    \centering
    \includegraphics[width=0.8\columnwidth]{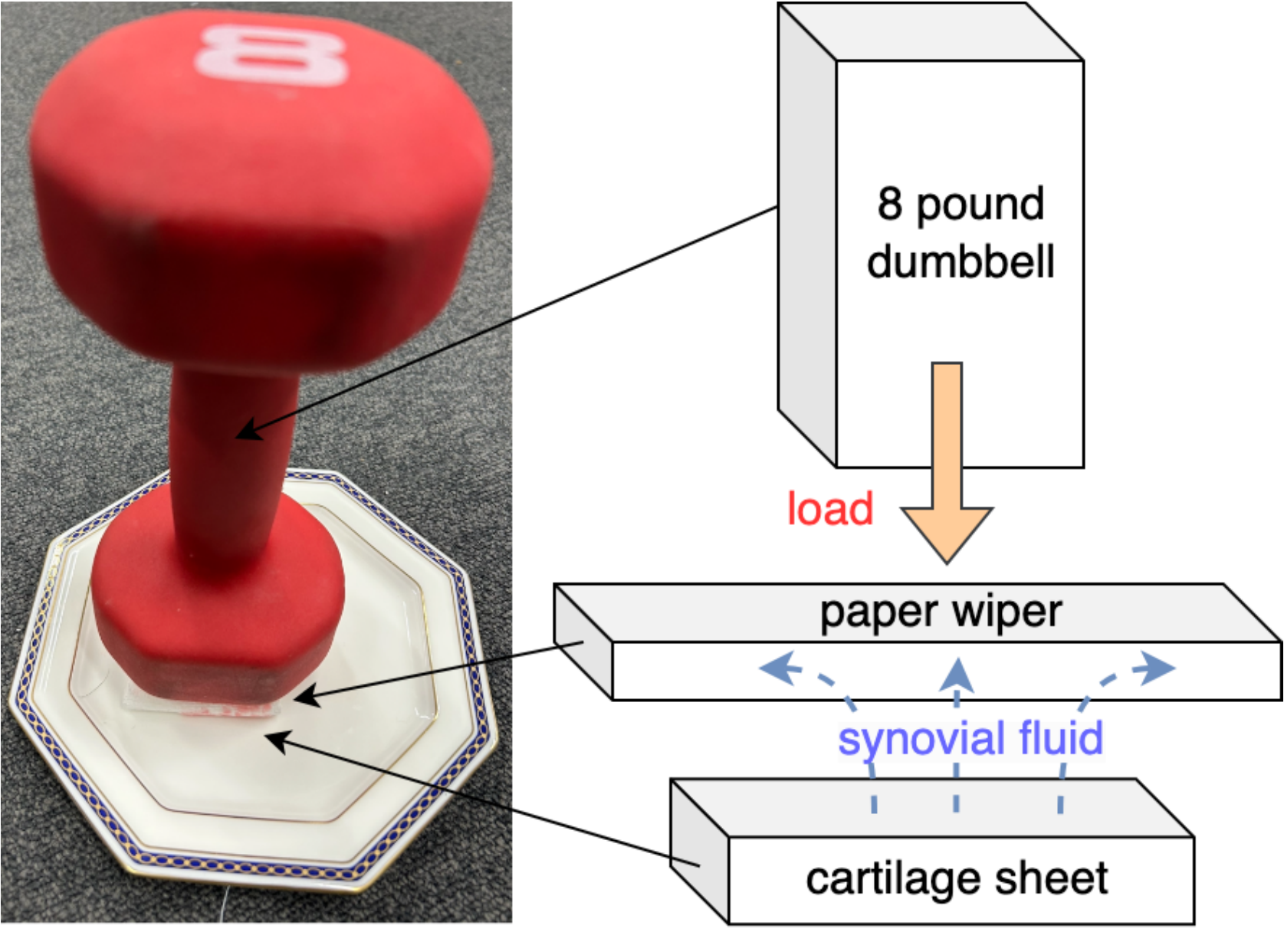}
    \caption{Overview of fluid exudation experiment under load conditions.}
    \label{fig:pre-experiment-overview}
  \end{center}
\end{figure}

\begin{figure}[t]
  \begin{tabular}{cc}
    \begin{minipage}[b]{.33\columnwidth}
      \centering
      \captionsetup{width=.9\columnwidth}
      \includegraphics[width=0.75\columnwidth]{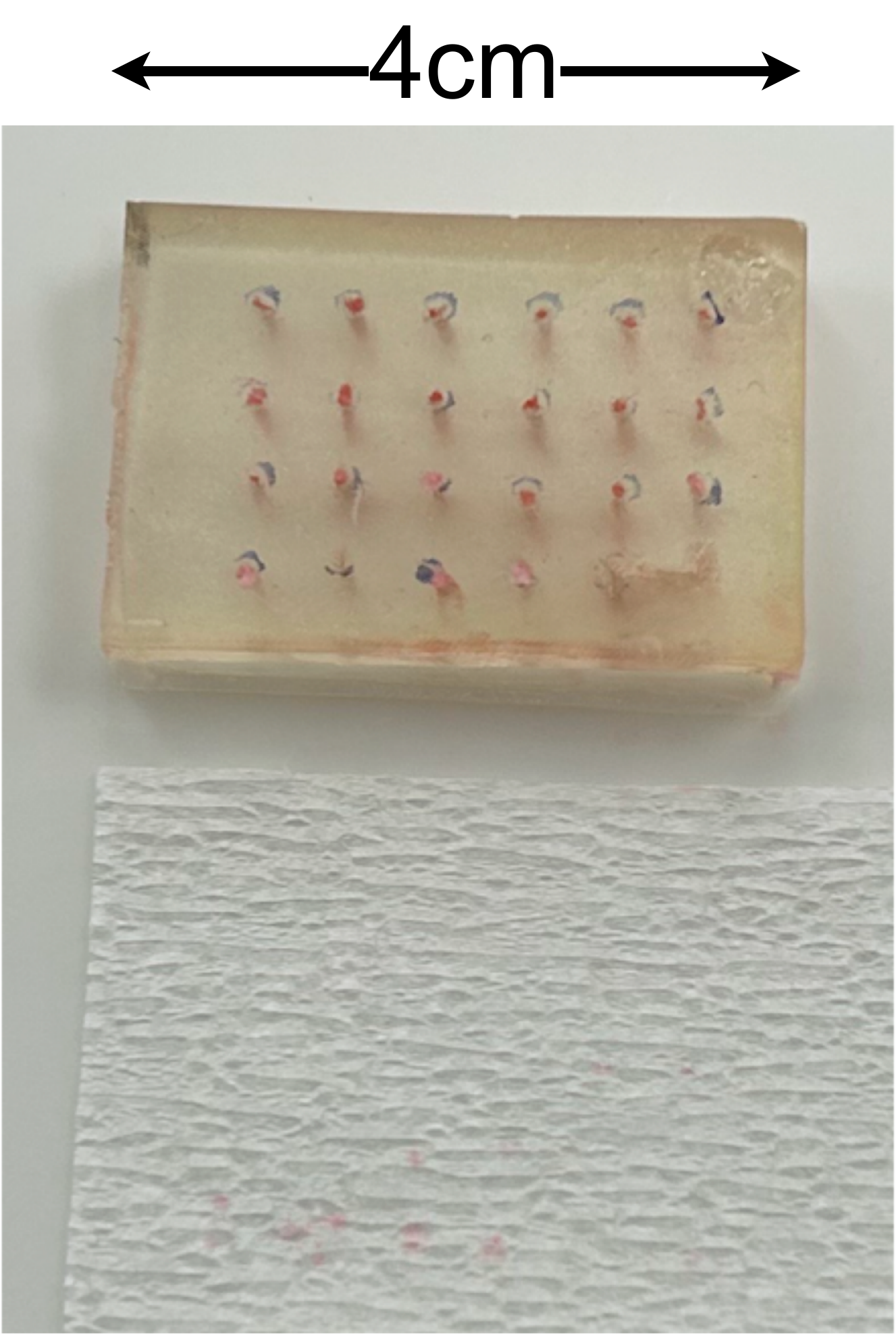}
      \caption{Fluid exudation experiment with \SI{0.5}{mm} diameter kite string (8-pound load).}
      \label{fig:pre-experiment-0.5mm}
    \end{minipage}
    \begin{minipage}[b]{.33\columnwidth}
      \centering
      \captionsetup{width=.9\columnwidth}
      \includegraphics[width=0.75\columnwidth]{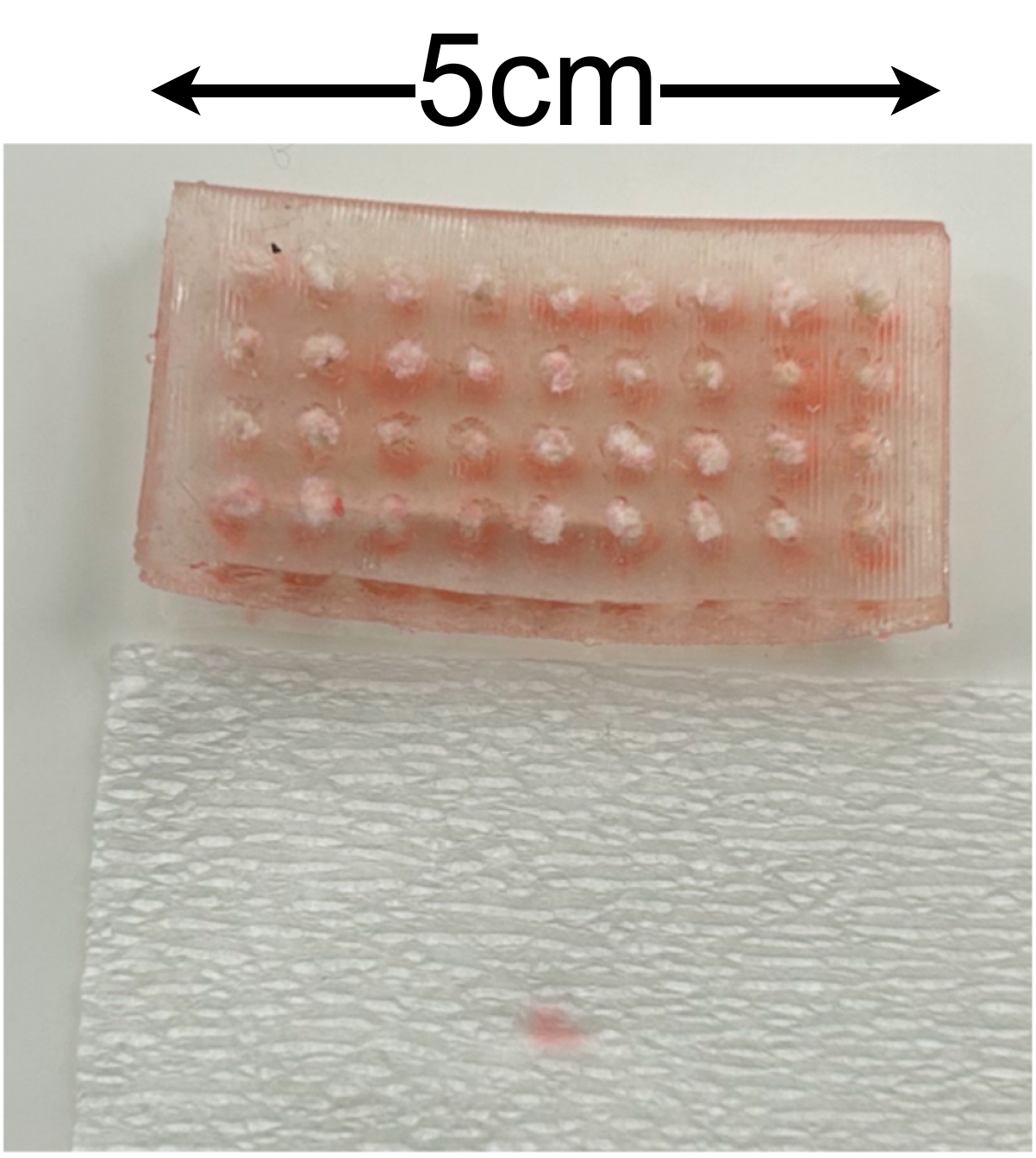}
      \caption{Fluid exudation experiment with \SI{1.6}{mm} diameter kite string (8-pound load).}
      \label{fig:pre-experiment-1.6mm}
    \end{minipage}
    \begin{minipage}[b]{.33\columnwidth}
      \captionsetup{width=.9\columnwidth}
      \centering
      \includegraphics[width=0.75\columnwidth]{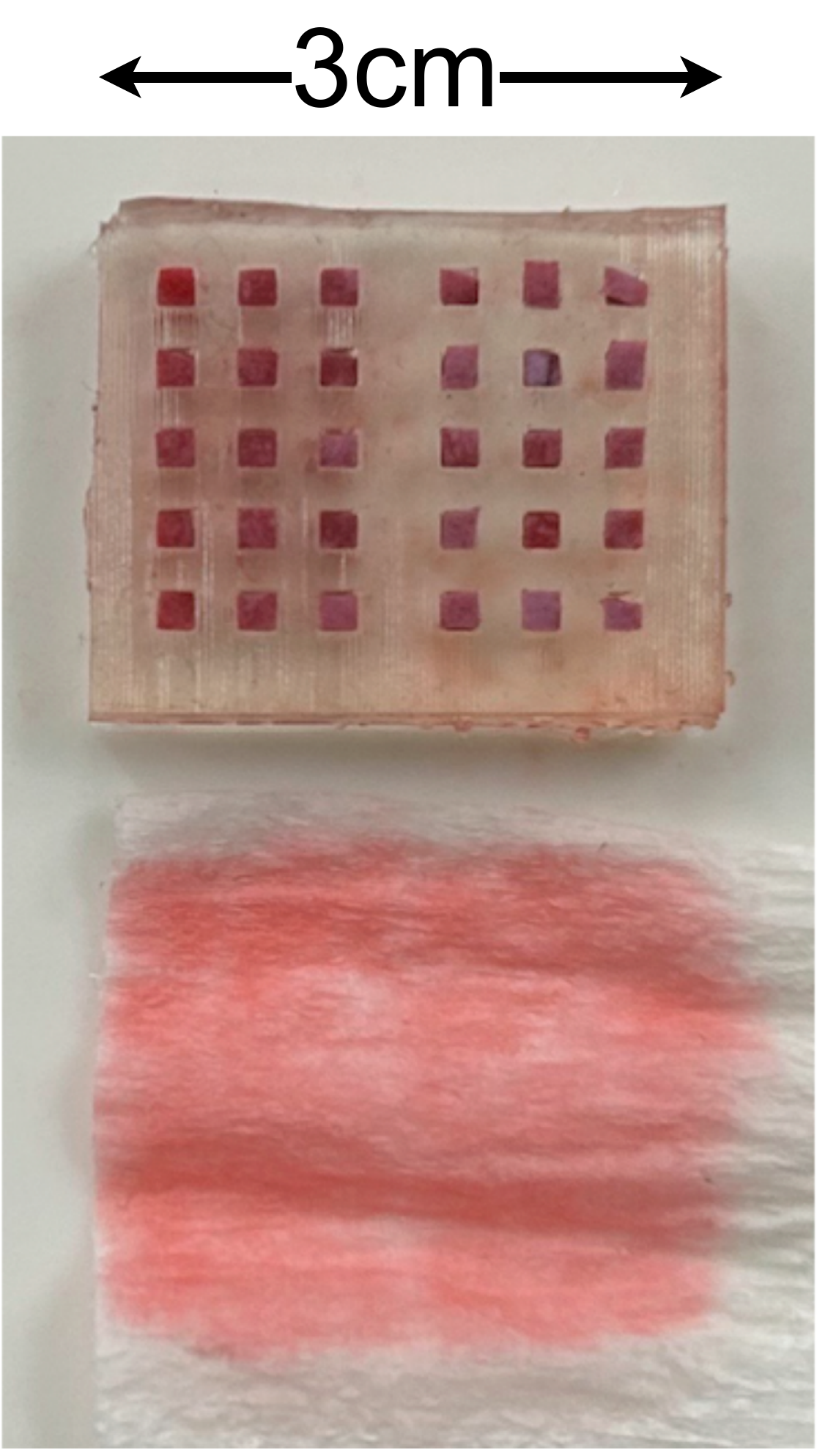}
      \caption{Fluid exudation experiment with PVA sponge (8-pound load).}
      \label{fig:pre-experiment-pva-8pound}
    \end{minipage}
  \end{tabular}
\end{figure}

スポンジを用いて作成した軟骨シートに関しては，荷重量に応じた滲出量の変化を確認する追加実験も行った．
8ポンドダンベル(\approx\SI{3.6}{kg})の荷重負荷をかけた後に紙ワイパーで滲出を吸収する操作を何度か繰り返すと，\figref{fig:pre-experiment-pva-8pound-little-exudation}のように紙ワイパーが吸い取る量は減少する．
吸い取り量が減少した\figref{fig:pre-experiment-pva-8pound-little-exudation}の状態から，15ポンドダンベル(\approx\SI{6.8}{kg})に荷重負荷を増大させる．
するとゴムの変形量が増えるため，液体の滲出量も増えると想定される．
結果は\figref{fig:pre-experiment-pva-15pound-more-exudation}のようになり，荷重量に応じて，さらなる液体滲出を確認できた．

\begin{figure}[t]
  \begin{tabular}{cc}
    \begin{minipage}[t]{.48\columnwidth}
      \centering
      \captionsetup{width=.9\columnwidth}
      \includegraphics[width=0.75\columnwidth]{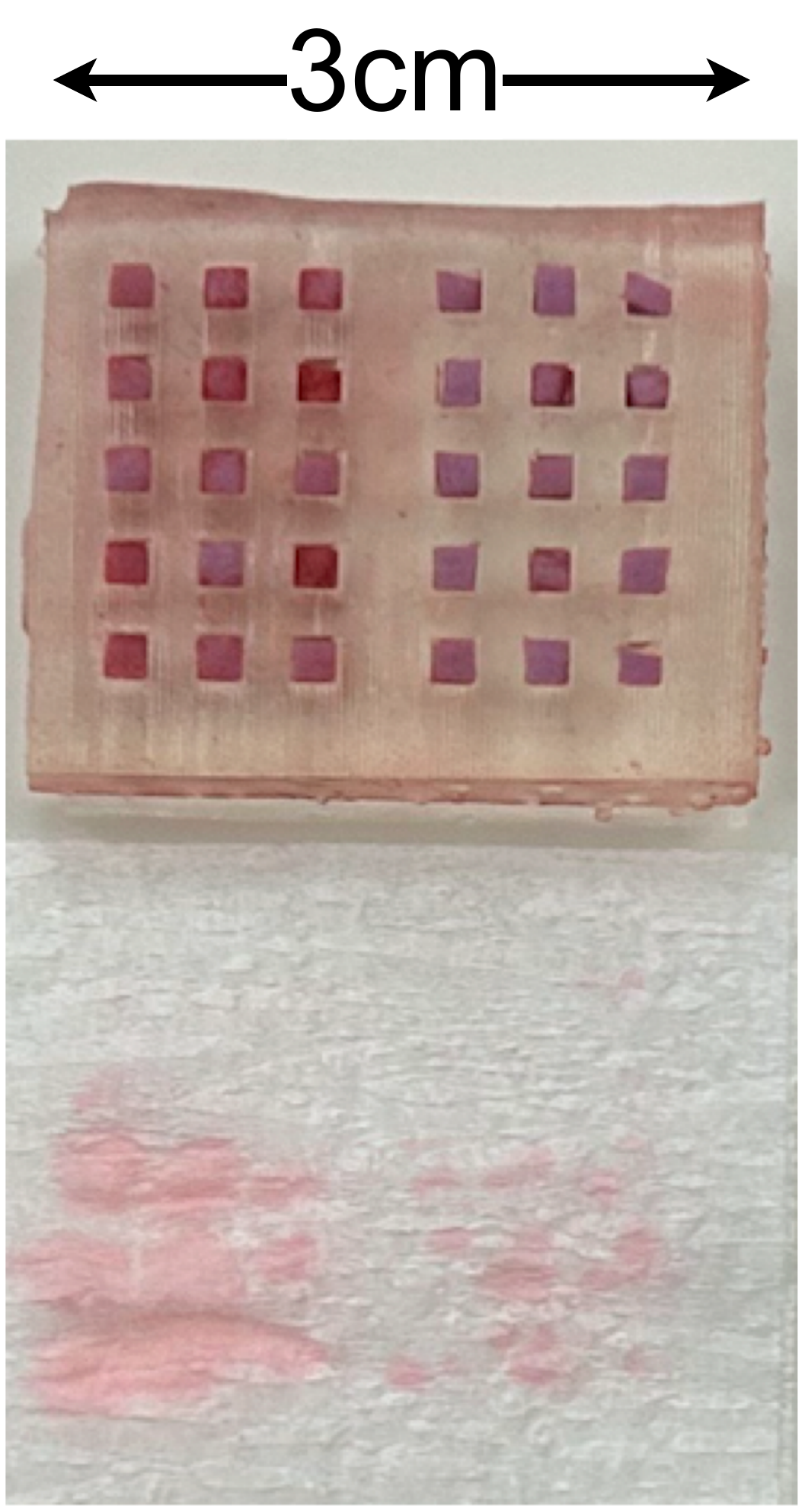}
      \caption{\textcolor{red}{Fluid exudation experiment with PVA sponge after many times (8-pound load).}}
      \label{fig:pre-experiment-pva-8pound-little-exudation}
    \end{minipage}
    \begin{minipage}[t]{.48\columnwidth}
      \centering
      \captionsetup{width=.9\columnwidth}
      \includegraphics[width=0.75\columnwidth]{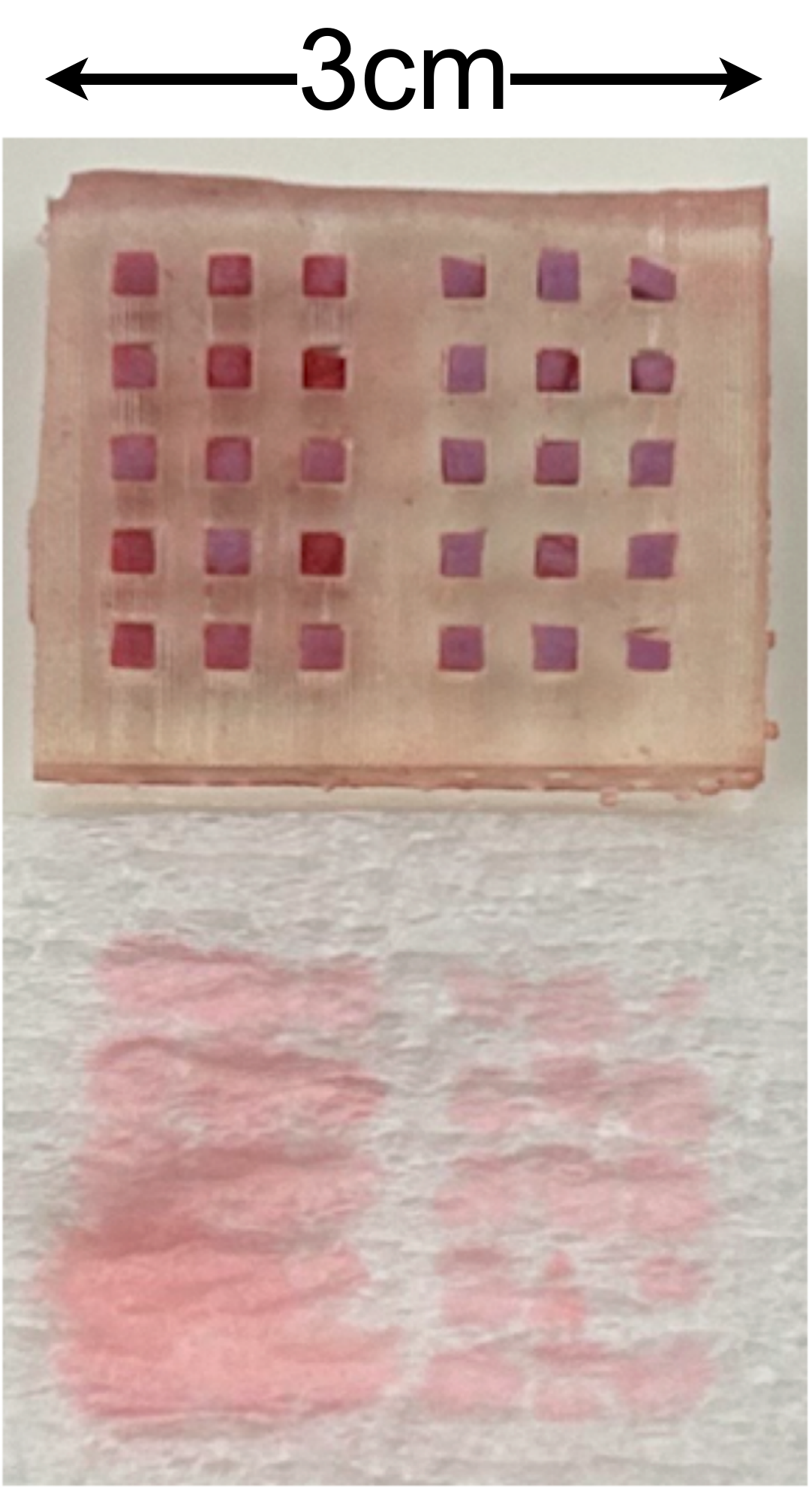}
      \caption{\textcolor{red}{Fluid exudation experiment with PVA sponge (15-pound load after \figref{fig:pre-experiment-pva-8pound-little-exudation}).}}
      \label{fig:pre-experiment-pva-15pound-more-exudation}
    \end{minipage}
  \end{tabular}
\end{figure}

さらに，軟骨シート下部にスポンジをシート状にひいた場合とひいていない場合で，荷重を繰り返しかけた際に液体滲出性能が保持されるかを比較検証した．
8ポンドダンベルの荷重負荷をかけた後に紙ワイパーで吸い取る操作を5回繰り返し，5回目操作後の紙ワイパーへの液体滲出量を比較した．
各操作の間では荷重無負荷の状態をはさみ，紙ワイパーを毎回交換した．
結果を\figref{fig:pre-experiment-pva-8pound-repeat-without-sheet}および\figref{fig:pre-experiment-pva-8pound-repeat-with-sheet}に示す．
軟骨シートの下にスポンジをシート状にひくことで液体滲出機能が保持された．
軟骨シート下部のスポンジシートが荷重負荷のないときに関節液を吸い上げ，軟骨シートに関節液を供給していることがわかる．
\begin{figure}[t]
  \begin{tabular}{cc}
    \begin{minipage}[b]{.48\columnwidth}
      \centering
      \captionsetup{width=.9\columnwidth}
      \includegraphics[width=0.75\columnwidth]{figs/pre_experiment_pva_8pound_5times.pdf}
      \caption{Fluid exudation experiment using PVA sponge after applying 8-pound load 5 times without base sheet.}
      \label{fig:pre-experiment-pva-8pound-repeat-without-sheet}
    \end{minipage}
    \begin{minipage}[b]{.48\columnwidth}
      \centering
      \captionsetup{width=.9\columnwidth}
      \includegraphics[width=0.75\columnwidth]{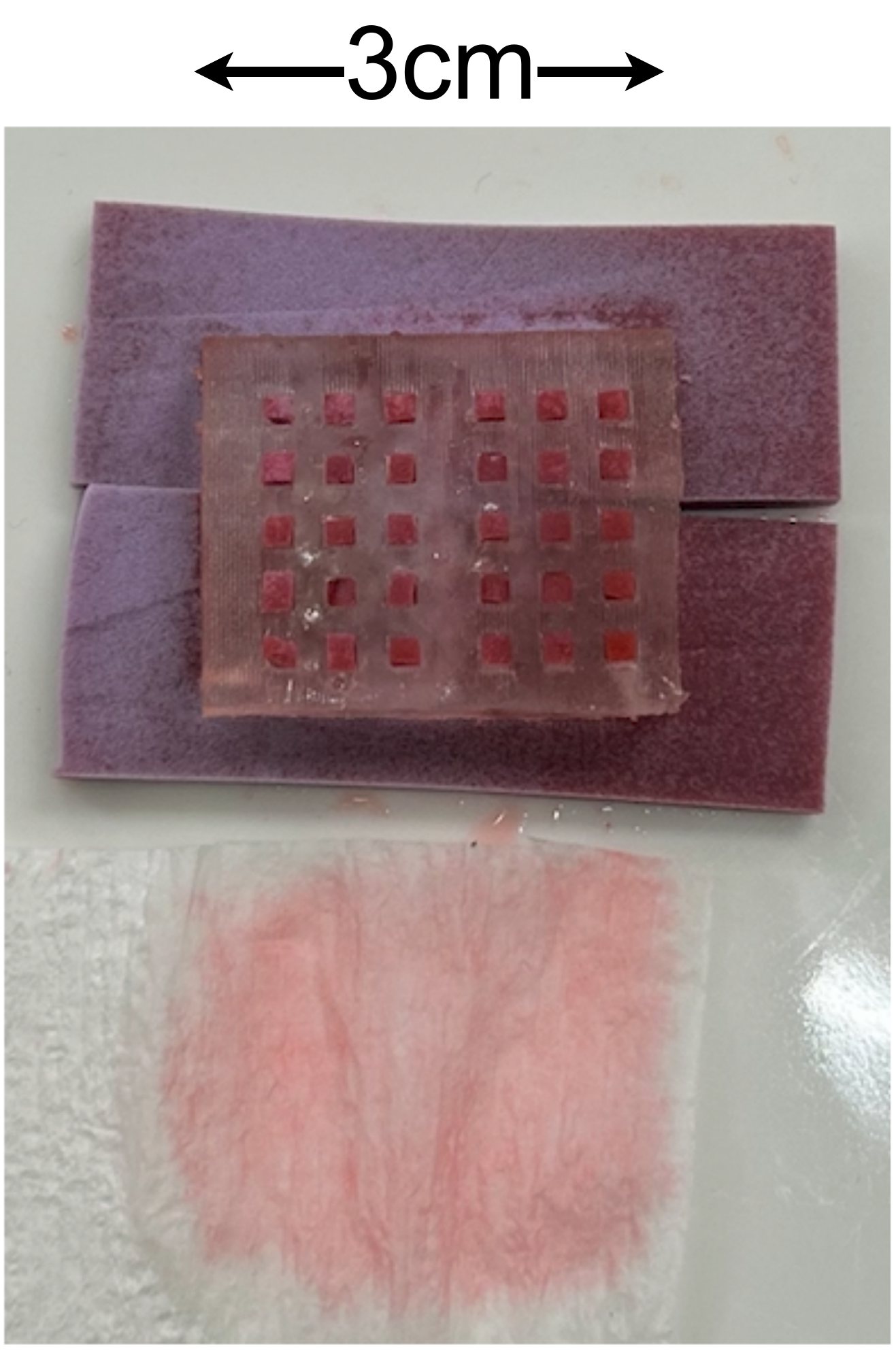}
      \caption{Fluid exudation experiment using PVA sponge after applying 8-pound load 5 times with base sheet.}
      \label{fig:pre-experiment-pva-8pound-repeat-with-sheet}
    \end{minipage}
  \end{tabular}
\end{figure}

これらの結果より，スポンジで作成した軟骨シートが，荷重負荷時に液体を滲出する機能を持つことを確認できた．
また，荷重量に応じて液体の滲出量が増えることを確認できるとともに，スポンジシートを軟骨シートの下部にひくことで液体滲出機能が保持されることも確認できた．

挿入する吸水性素材のサイズについては，生体軟骨内部の微細構造を考慮すると，小さなサイズを大量に挿入することが望ましい．
手加工可能なサイズとして，\SI{2}{mm}平方で吸水性素材を挿入することとした．


\textcolor{red}{
軟骨の曲率を考えることで，荷重負荷によって軟骨シートが圧縮変形した際に，常に吸水性素材挿入部が荷重面に接するように設計することができる．
三角関数を用いた計算により，吸水性素材挿入部は\SI{2}{mm}間隔で配置することで十分である．
以下，詳細に説明する．
}

\textcolor{red}{
軟骨の曲率を$1/r$とすると，軟骨表面は半径$r$の仮想円を形作る．
荷重負荷によって，\figref{fig:cartilage-surface-under-load}のように$\delta$ mm 弾性変形したとする．
}
\begin{figure}[t]
  \begin{center}
    \centering
    \includegraphics[width=0.8\columnwidth]{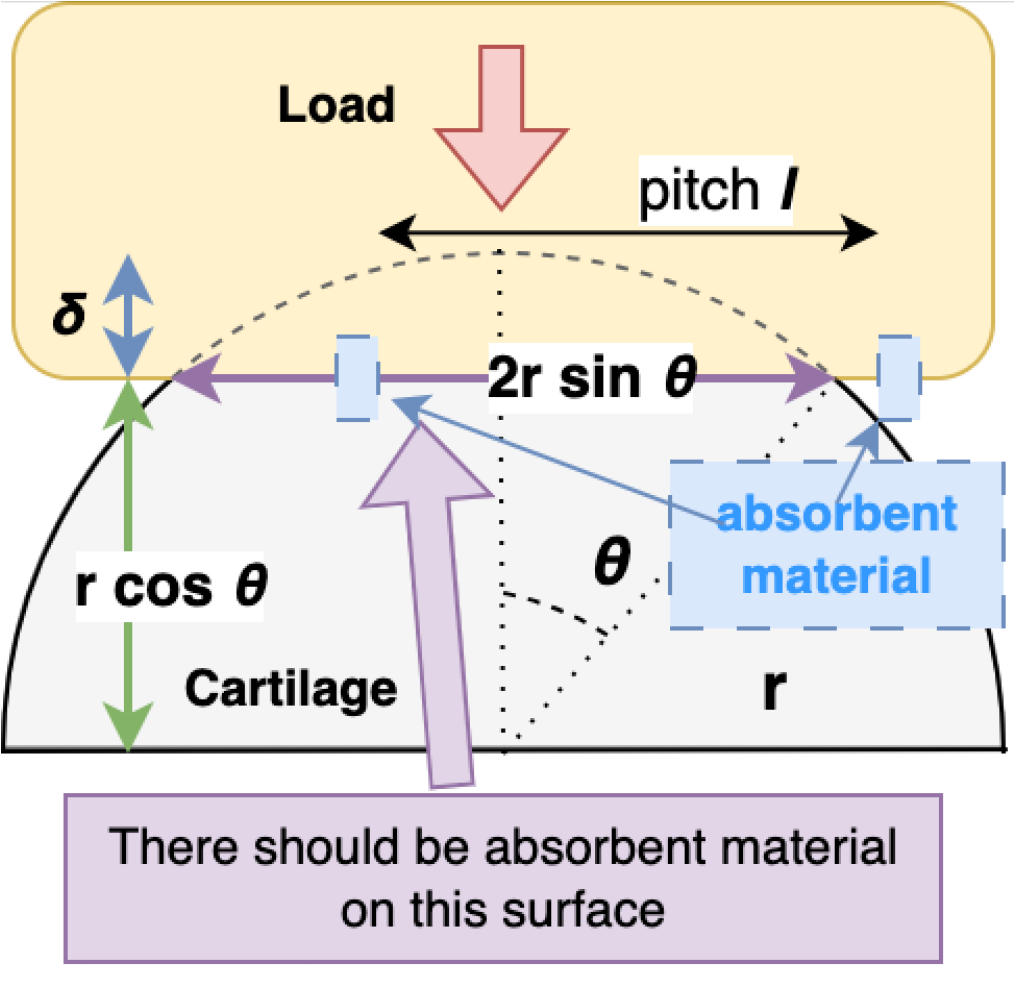}
    \caption{Figure of cartilage surface under load conditions.}
    \label{fig:cartilage-surface-under-load}
  \end{center}
\end{figure}
\textcolor{red}{
このとき，吸水性素材挿入部のピッチ間隔を$l$ mm とすると，弾性変形した際に形成される平面に常に吸水性素材挿入部が存在するためには，以下の式が成り立てば良い．
\begin{align}
  r\cos(\theta) &= r - \delta \\
  2r\sin(\theta) &\ge l \label{eq:pitch-condition}
\end{align}
三角関数の関係から\equref{eq:pitch-condition}は以下のように変形できる．
\begin{align}
  l &\le 2r\sin(\theta) \notag \\
    &\le 2r\sqrt{1 - (\cos(\theta))^2} \notag \\
    &\le 2\sqrt{r^2 - (r-\delta)^2} \notag \\
    &\le 2\sqrt{\delta(2r-\delta)} \label{eq:pitch}
\end{align}
現実的に，液体浸潤軟骨が効果を発揮するような高荷重状態における軟骨シートのたわみ量$\delta$は，\SI{1}{mm}に近いものとなる．
実際に\SI{2}{mm}平方の吸水性素材挿入部を約\SI{2}{mm}間隔で配置した軟骨シートにおいて，15ポンドダンベル(\approx\SI{6.8}{kg})の重りを乗せた際には約\SI{1}{mm}圧縮変形した．
\textcolor{red}{
これは人間の身体の日常的な動作を考えても，十分にありうる荷重であるといえる．
}
吸水性素材挿入部を配置しない軟骨シートにおいても，15ポンドダンベル(\approx\SI{6.8}{kg})の重りを乗せた際には，配置した場合に比べて，0.1$\sim$\SI{0.2}{mm}程度の差しか見られなかった．
また，軟骨の曲率に関しても，人間の骨の大きさを考慮すると，$r\geq\SI{3}{mm}$と近似できる．
その場合，
\textcolor{red}{
荷重条件下では，
}
\equref{eq:pitch}における根号の中は1以上となるので，$l<=\SI{2}{mm}$であれば，弾性変形した際に形成される平面に常に吸水性素材挿入部が存在するための条件は満たされる．
上記の議論より，以後の実験では，\SI{2}{mm}平方の吸水性素材挿入部を約\SI{2}{mm}間隔で配置した軟骨シートを用いる．
}


\textcolor{red}{
  \SI{2}{mm}平方の吸水性素材挿入部を約\SI{2}{mm}間隔で配置した軟骨シートに関して，荷重負荷時のたわみ量についても考察を行った．
  Elastic50Aのヤング率を$E$ [\si{kgf/cm^2}]，ポアソン比を$\nu$，静的せん断弾性率を$G$ [\si{kgf/cm^2}]とする．
  また，荷重を$W$ [\si{kgf}]，見かけのヤング率を$E_{ap}$ [\si{kgf/cm^2}]，荷重を受ける総面積を$A_L$ [\si{cm^2}]，ゴムの高さを$h$ [\si{cm}]，ゴムのたわみ量を$\delta$ [\si{cm}]とする．
  ただし，弾性変形は荷重負荷面が十分大きな平面であると近似し，軟骨シートの最端部の変形は考慮しないものとする．
  また，吸水性素材がゴムより十分柔らかいと仮定する．
  この時，\equref{eq:gomu-tawami}が成り立つことが知られる．
  \begin{equation}
    \delta = \frac{W}{E_{ap}\times{A_L/h}} \label{eq:gomu-tawami}
  \end{equation}
  荷重を受けた場合に変形可能な部分の総面積である自由面積を$A_F$ [si{cm^2}]とすると，形状率$S$は\equref{eq:shape-rate}で定義される．
  \begin{equation}
    S = \frac{A_L}{A_F} \label{eq:shape-rate}
  \end{equation}
  四角柱がくり抜かれる形のゴムの角柱において，見かけのヤング率$E_{ap}$は，
  \begin{equation}
    E_{ap} = G (4+3.290S^2) \label{eq:pseudo-young-modulus}
  \end{equation}
  と表現されることが知られている．
  本研究で作成した軟骨シートは穴空き型の角柱なので，\figref{deformation-cartilage-under-load}のように単位角柱を定めると，
  \begin{equation}
    S = \frac{a^2-b^2}{4bh} \label{eq:shape-rate-detail}
  \end{equation}
  と表現される．
}

\begin{figure}[t]
  \begin{center}
    \centering
    \includegraphics[width=\columnwidth]{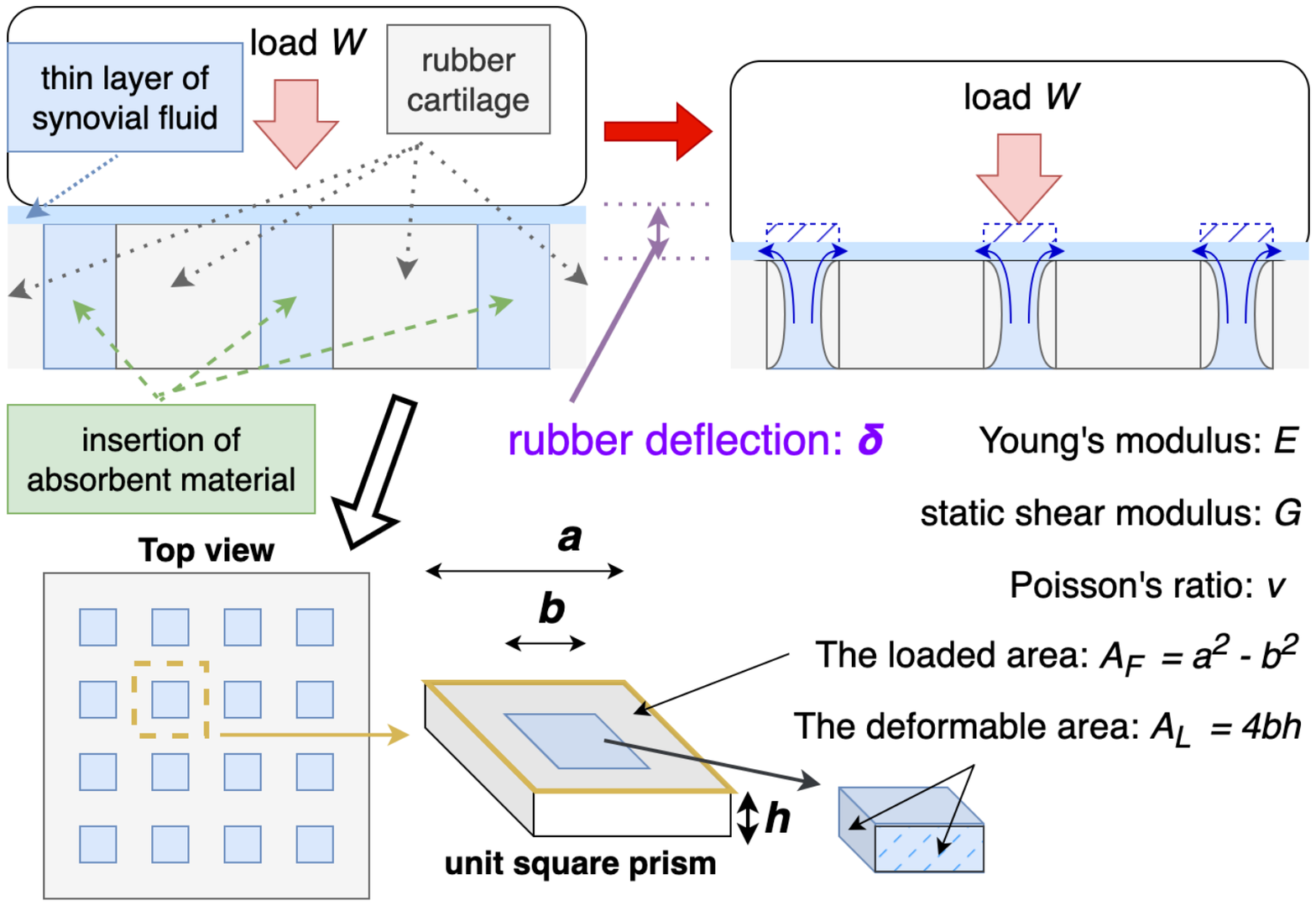}
    \caption{Figure of elastic deformation of cartilage sheet under load.}
    \label{fig:deformation-cartilage-under-load}
  \end{center}
\end{figure}

\textcolor{red}{
  軟骨シートを厚さ\SI{0.6}{cm}で作成した場合，$a=\SI{0.4}{cm}$, $b=\SI{0.2}{cm}$, $h=\SI{0.6}{cm}$である．
  このとき，\equref{eq:shape-rate-detail}より$S=0.25$であるから，見かけのヤング率$E_{ap}$は\equref{eq:pseudo-young-modulus}から，$E_{ap}\approx\SI{4.2}{kgf/cm^2}$となる．
  さらに，一般的な弾塑性体モデルの多くで，\equref{eq:G}が成り立つ．
  \begin{equation}
    G = \frac{E}{2(1+\nu)} \label{eq:G}
  \end{equation}
  よって，Elastic50Aのヤング率は$E\approx\SI{3}{kgf/cm^2}$で，ポアソン比は$\nu=0.5$とみなせることから，$G\approx\SI{1}{kgf/cm^2}$であるといえる．
  例えば，単位角柱9個分が連なった正方形形の軟骨シートに15ポンドダンベル(\approx\SI{6.8}{kg})をのせたときのたわみ量は，一つあたりの単位角柱にかかる荷重は$1/9$になるので，\equref{eq:gomu-tawami}より
  \begin{align}
    \delta &= \frac{W}{E\_{ap}\times{A_L/h}} \label{eq:4} \\
           &\approx \frac{6.8/9}{4.2 \times (0.12/0.6)} \\
           &\approx 0.9
  \end{align}
  となる．
  実際に，15ポンドダンベル(\approx\SI{6.8}{kg})の重りを単位角柱9個分の軟骨シートに乗せた際に約\SI{1}{mm}圧縮変形した．
  これは計測結果と一致しており，上記の議論が概ね正しいことを示している．
}

\textcolor{red}{
  ここまでの議論をもとにして，スポンジが圧縮される体積を考えることで，液体滲出量を考察する．
  荷重方向に対して，スポンジが圧縮される体積量$L_1$は\equref{eq:L1}である．
  \begin{equation}
    L_1 = b^2 \delta \label{eq:L1}
  \end{equation}
  側面方向からスポンジが圧縮される体積量$L_2$はゴムの変形量に対してポアソン比を用いて\equref{eq:L2}となる．
  \begin{equation}
    L_2 = \nu(a^2-b^2)\delta  \label{eq:L2}
  \end{equation}
  結局，スポンジが圧縮される総体積量は，\equref{eq:L-total}のようになる．
  \begin{align}
    L_{total} &= L_1 + L_2 \notag \\
      &= (b^2 + \nu a^2 - \nu b^2) \delta \notag \\
      &= (\nu a^2 + b^2 - \nu b^2) \times \frac{W}{E\_{ap}\times{A_L/h}} \label{eq:L-total}
  \end{align}
  特に，$a=\SI{0.4}{cm}$，$b=\SI{0.2}{cm}$，$h=\SI{0.6}{cm}$のパッキン型の単位角柱が$x$個連続した軟骨シートに対して，荷重$W$ [\si{kgf}]をかけた場合，$\nu\approx0.5$，$A_L = \SI{0.12}{cm^2}$，$E\_{ap}\approx\SI{4.2}{kgf/cm^2}$であるので，
  \begin{equation}
  L_{total} \approx \frac{17W}{21x} \label{eq:L-total-numerical}
\end{equation}
  となる．
  実際には，\equref{eq:L-total-numerical}で求めた吸水性素材の圧縮体積量に対応して関節液が滲出する．
}

}%

\section{Main Experiment: Friction Coefficient Measurement Experiment}
\label{sec:main-experiment}
\switchlanguage%
{%
  We constructed a friction testing machine, as shown in \figref{fig:friction-test-overview}, and conducted experiments to measure the coefficient of friction.
  Creating an actual open-type ball joint to measure the friction coefficient is challenging due to the significant influence of joint capsules and ligaments, and the difficulty in measuring friction on a spherical surface.
  Therefore, we measured the friction coefficient on a planar sheet to conduct a fundamental evaluation.

  Various methods have been developed for friction testing, one of which is the pin-on-plate test.
  The pin-on-plate test is a widely used method
  and has also been used in recent years for evaluating the friction and wear characteristics of artificial joint materials\cite{yarimitsu2016PVA-hydrogel}.
  For example, in testing artificial joint materials, a plate-shaped cartilage piece is fixed to a base, and a pin-shaped articular cartilage is brought into contact with it.
  The two are then slid against each other by applying a load and either rotating or translating them using a motor.
  The average friction force is measured using a load cell connected to the base or the pin, and the friction coefficient is calculated.
  However, in the pin-on-plate test machine, there are issues such as the fact that high loads are constantly applied to the pin test piece during rotation or translation, and that precision testing machines are very expensive.
  In human and robotic movements, as seen in bending and push-up exercises, high and low loads are repeated, and high load conditions are not always sustained.
  In fact, it has been reported that the friction coefficient increases as the high load conditions persist in human cartilage\cite{forster1999influence}, indicating that it is not a suitable condition for use.
  As the cartilage sheet developed in this study is intended for use in bio-inspired robots, we constructed a friction testing machine and experiments were conducted while avoiding continuous high-load conditions.

  The procedure for this experiment is as follows.
  We used a pseudo-synovial fluid, diluted hyaluronic acid solution to 3-\SI{4}{mg/ml}.
  We placed the cartilage sheet inside an acrylic case capable of holding synovial fluid, and placed weights on top of it.
  The weights were fixed dumbbells on an aluminum frame, with the bottom surface of the aluminum frame resting on the cartilage sheet.
  A large pulley was attached to the acrylic case.
  One side of the large pulley was connected to a weight using a string, and the other side had a suspended PET bottle filled with water.
  When the weight-loaded aluminum frame is slowly placed on the cartilage sheet, it is checked whether it slips due to being pulled by the large pulley.
  If it does not slip, the aluminum frame is removed from the cartilage sheet once, the water in the PET bottle is increased, and it is placed again.
  By repeating the above operation, we determined the weight $W$ at which the PET bottle began to slip and the relationship $W = \mu N$ between the weight on the cartilage sheet $N$ and the weight of the slipping PET bottle $W$, thus obtaining the coefficient of friction $\mu$.
  A low-friction PTFE film was affixed to the bottom surface of the aluminum frame, and we examined different conditions such as attaching a low-friction PTFE film to the friction surface of the planar cartilage sheet, immersing the load surface in synovial fluid, and introducing a fluid exudation mechanism.
  An 8-pound dumbbell ($\approx$\SI{3.6}{kg}) was used for the load.
  By conducting friction tests in this way, friction tests were conducted while avoiding continuous high-load conditions.

\begin{figure}[t]
  \begin{center}
    \includegraphics[width=0.9\linewidth]{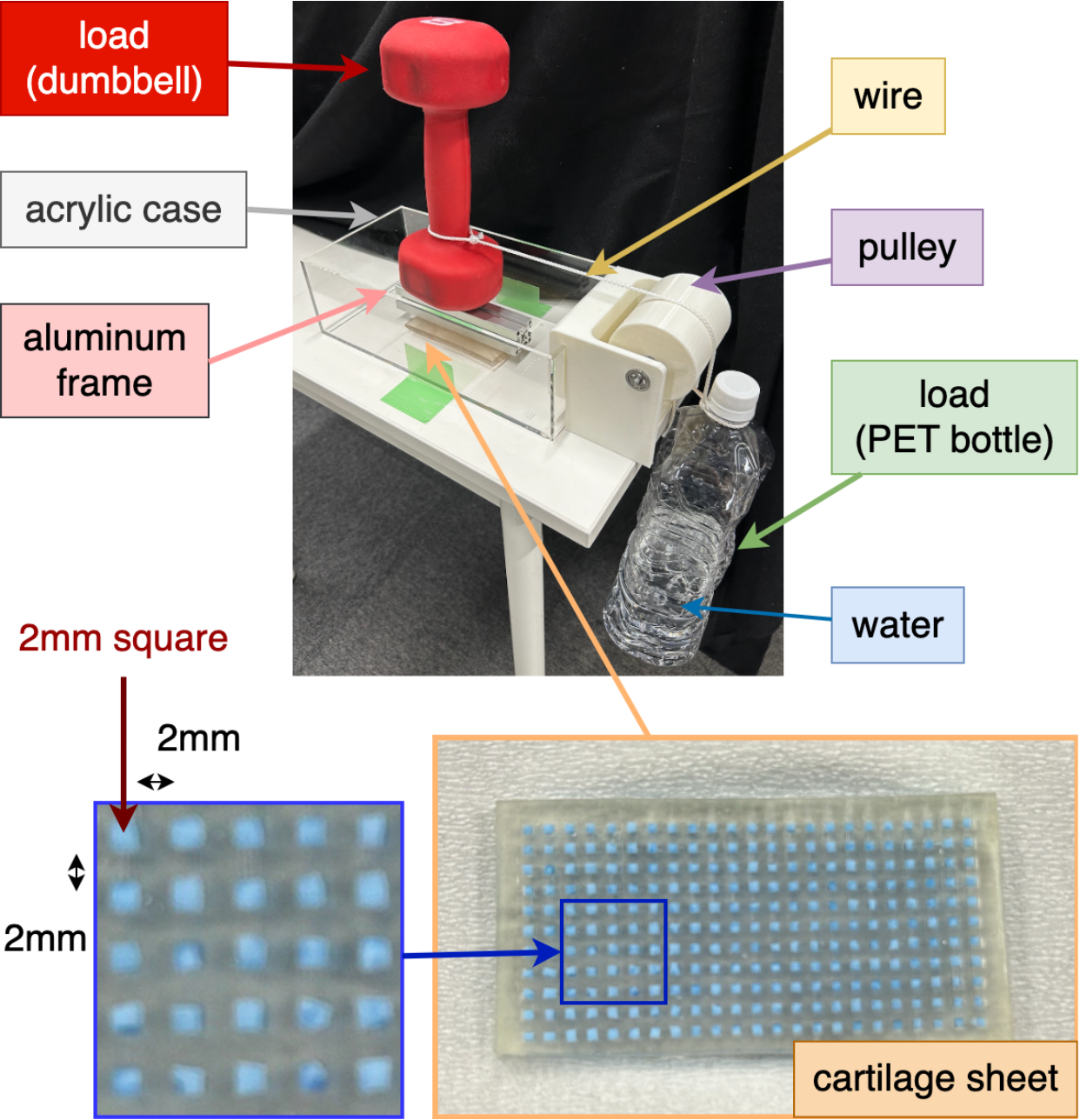}
    \caption{Overview of friction coefficient measurement experiment.}
    \label{fig:friction-test-overview}
  \end{center}
\end{figure}

  The results are presented in \tabref{tab:friction-coefficient}.
  In the state of rubber alone without a film, it did not move even with a \SI{2}{L} PET bottle.
  The coefficient of friction is found to be above 0.5, which is a typical result as the coefficient of friction for rubber is generally considered to be between 1 and 3.
  Since coefficients of friction of this magnitude are not suitable for use, further measurements were not conducted.
  Focusing on the conditions with the PTFE film, it can be observed that the coefficient of friction on the surface in contact with the load is reduced.
  Moreover, not only the use of the film but also immersing the load surface in synovial fluid led to a further reduction in the coefficient of friction.
  It was evident that fluid lubrication was effective.
  Furthermore, by introducing a fluid exudation mechanism, the coefficient of friction was further reduced.
  Compared to simply immersing the load surface in synovial fluid, the introduction of the fluid exudation mechanism demonstrated a more uniform fluid lubrication to the surface in contact with the load.
  Additionally, in the case of simply immersing the load surface in synovial fluid, a significant deterioration in sliding performance was observed after holding the load in the same position for a constant period.
  It is believed that, when held for a certain period without the fluid exudation mechanism, synovial fluid is pushed out from the load contact surface, resulting in solid-solid contact.
  In contrast, when the fluid exudation mechanism was introduced, a significant deterioration in sliding performance was not observed even after holding the load in the same position for the same amount of time.

  From the results of this experiment, it was confirmed that the fluid exudation function is essential for reducing the coefficient of friction.
  However, it was not possible to reduce the coefficient of friction to the order of 0.005, equivalent to that of bearings.
  This is likely because the proposed fluid-exuding cartilage in this study does not perform fluid exudation with the same microstructure as living cartilage.
  Nevertheless, it is undoubtedly lower friction than solid contact friction, and it can also be considered that the fluid-exuding cartilage in this study somewhat achieves the fine structure of biological cartilage in a macroscopic sense.
  Considering that the coefficient of friction of ultra-high molecular weight polyethylene used in artificial joints is between 0.11 and 0.17, a coefficient of friction value of 0.053 is less than half.
  This value places the friction coefficient in the order comparable to that of industrial sliding bearings.
  From this experiment, the usefulness of the proposed fluid-exuding cartilage mechanism was confirmed.

\begin{table}[t]
  \caption{Table of friction coefficient}
  \label{tab:friction-coefficient}
  \centering
  \begin{tabular}{cccr}
    \hline
    Measuring conditions & N [g] & W [g] & coefficient \\
    \hline \hline
    no film & 3843 & $\geq2000$ & $\geq0.5$ \\
    film & 3843 & 304 & 0.079 \\
    film + fluid & 3843 & 247 & 0.064 \\
    film + fluid + sponge & 3843 & 207 & 0.053 \\
    \hline
  \end{tabular}
\end{table}
}%

{%
}%

\section{Construction of an Open-Type Ball Joint Utilizing fluid-exuding Cartilage}
\label{sec:creation-open-type-ball-joint}
\switchlanguage%
{%
  We fabricated a curved fluid-exuding cartilage and integrated it with 3D-printed bones, ligaments crafted from natural rubber, synovial fluid, and a joint capsule to actualize an open-type ball joint, utilizing fluid-exuding cartilage, as illustrated in \figref{fig:construction-open-type-ball-joint}.
  Leveraging the advantages of 3D-printable rubber material, a curvature resembling authentic cartilage was achieved.
  Approximately \SI{2}{mm} square PVA sponges were inserted at intervals of about \SI{2}{mm} to create curved fluid-exuding cartilage.

  The base layer of PVA sponge was tightly attached to the joint contact surface of the formed bone and bonded with instant adhesive.
  Subsequently, the bone, the basal layer affixed to it, and the cartilage section above the basal layer were enveloped in films.
  The film was perforated with numerous holes to allow fluid exudation.

  We processed both the bone acting as the head and the bone serving as the socket following the steps mentioned above, and then fitted them together.
  Additionally, TPU film was used as the joint capsule, enveloping the entire joint and filling the interior with synovial fluid.
  Ligaments were employed from the outside of the joint capsule to connect the two bones, using natural rubber sheets.
  The Ligaments were fixed to the bones with insert nuts.
  The open ball-type joint constructed in this manner was confirmed to move smoothly with repeated manual manipulation, as shown in \figref{fig:range-of-motion}.
  While not achieving the extensive range of motion seen in natural human joints due to robust constraints imposed by the ligaments and resistance from the TPU film in the joint capsule, it repetitively executed rotational and twisting movements with smooth operation.

  Moving forward, further research is necessary to achieve a broader range of motion, particularly focusing on ligaments and joint capsules.

\begin{figure}[t]
  \begin{center}
    \includegraphics[width=\linewidth]{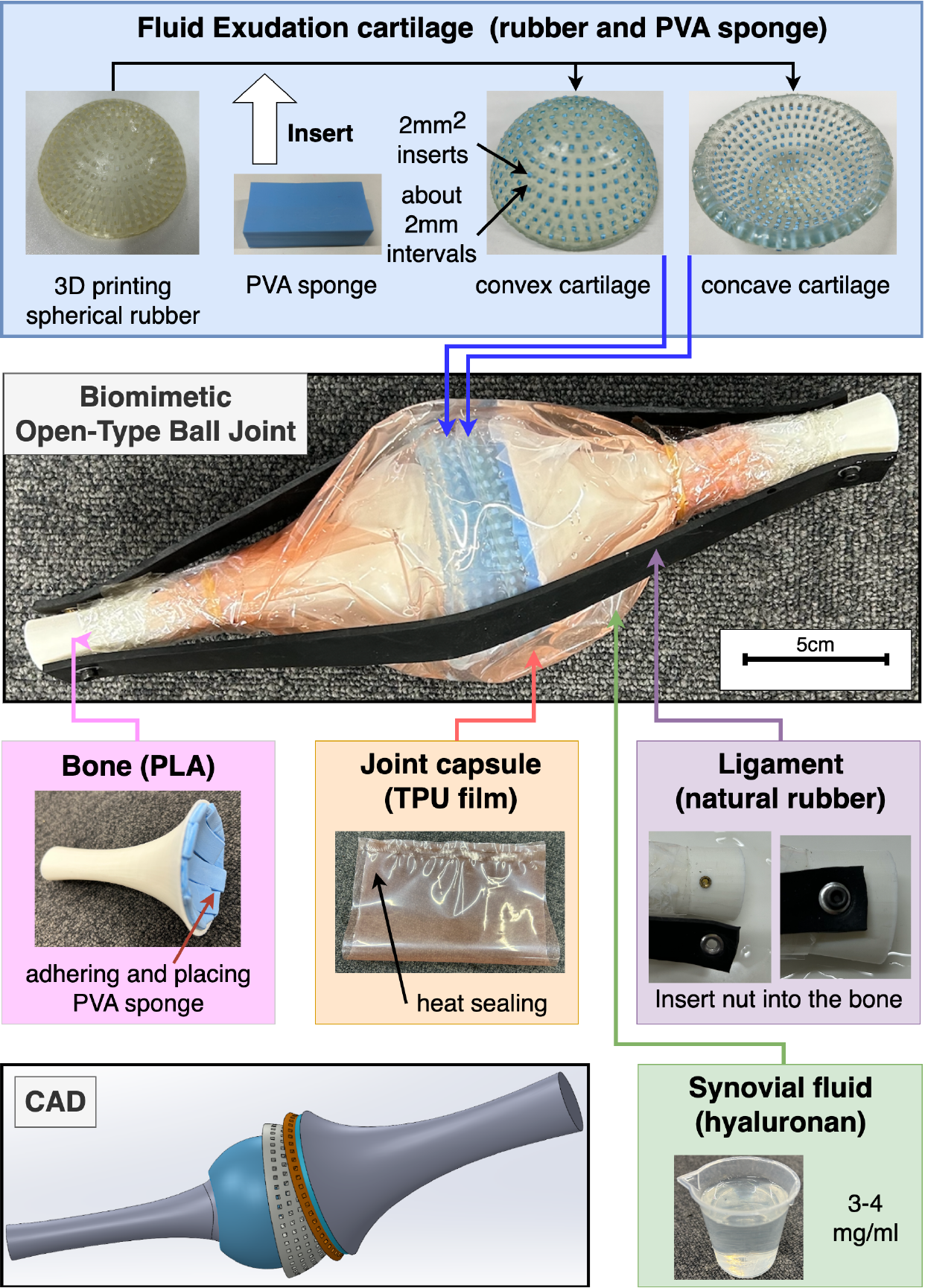}
    \caption{Figure illustrating the construction method of an open-type ball joint utilizing fluid exudation cartilage.}
    \label{fig:construction-open-type-ball-joint}
  \end{center}
\end{figure}

\begin{figure}[t]
  \begin{center}
    \includegraphics[width=\linewidth]{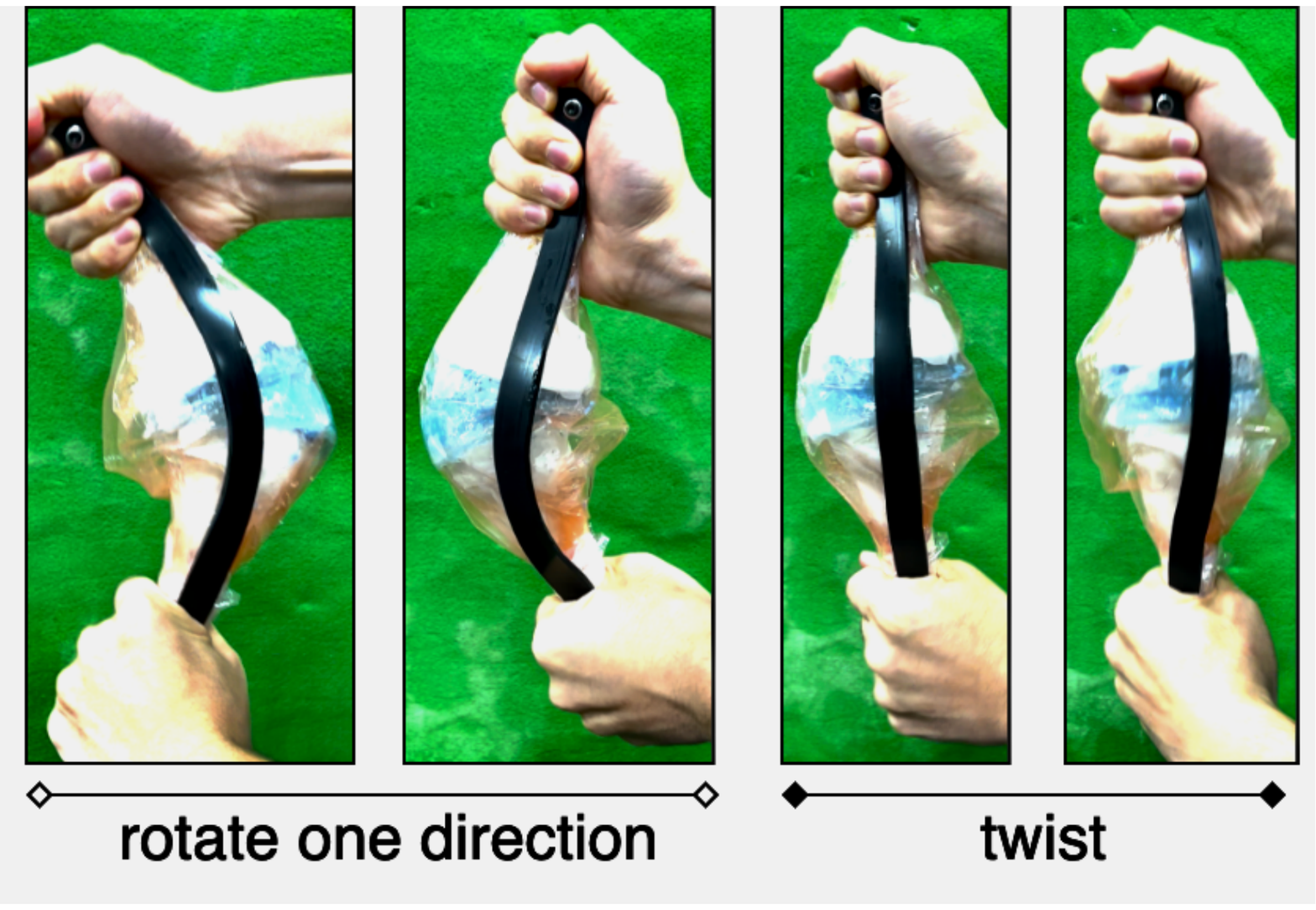}
    \caption{Figure illustrating the manual movement of an open-type ball joint.}
    \label{fig:range-of-motion}
  \end{center}
\end{figure}

}%
{%
曲面状の液体滲出軟骨を作成し，3Dプリンタで造形した疑似骨，天然ゴムによる模擬靭帯，関節液，関節包を組み合わせて実際に\figref{construction-open-type-ball-joint}のような液体滲出軟骨を用いた開放型球関節を構成した．
3Dプリンティングできるゴム素材の利点を活かして，実際の軟骨に近い曲面形状を造形し，約\SI{2}{mm}平方のPVAスポンジを約\SI{2}{mm}間隔で挿入し，曲面状の液体浸潤軟骨を作成した．
基底層のPVAスポンジは，造形した擬似骨の関節接触面にしきつめて，瞬間接着剤で接着できる．
そのうえ，擬似骨，擬似骨に接着された基底層，基底層の上にある軟骨部をフィルムで覆い，フィルムには液体が滲出できるように無数の穴をあけておく．
骨頭となる側の擬似骨と受け皿となる擬似骨の両方で，上記の処理を行った上で，はめあわせた．
さらに関節包としてTPUフィルムを用いて，関節全体を包み，関節液で内部を満たした．
関節包の外から模擬靭帯を用いて，2つの骨をつなぎ合わせる．
模擬靭帯には天然ゴムシートを用いて，インサートナットにより模擬骨に固定した．
このようにして構成した開放型球関節は，実際に手で動かすことで，繰り返しなめらかに動くことが確認できた．
\textcolor{red}{
実際に手で動かしている様子を\figref{fig:range-of-motion}に示す．
靭帯による強い拘束と関節包に用いたTPUフィルムの抵抗によって，人間の関節ほど大きな可動域をとることはできていないが，回転やひねりといった動きを繰り返し行うことができ，動作の過程もなめらかであった．
今後は，広い可動域を実現するために，靭帯や関節包に関して研究を進めていく必要がある．
}

\begin{figure}[t]
  \begin{center}
    \includegraphics[width=\linewidth]{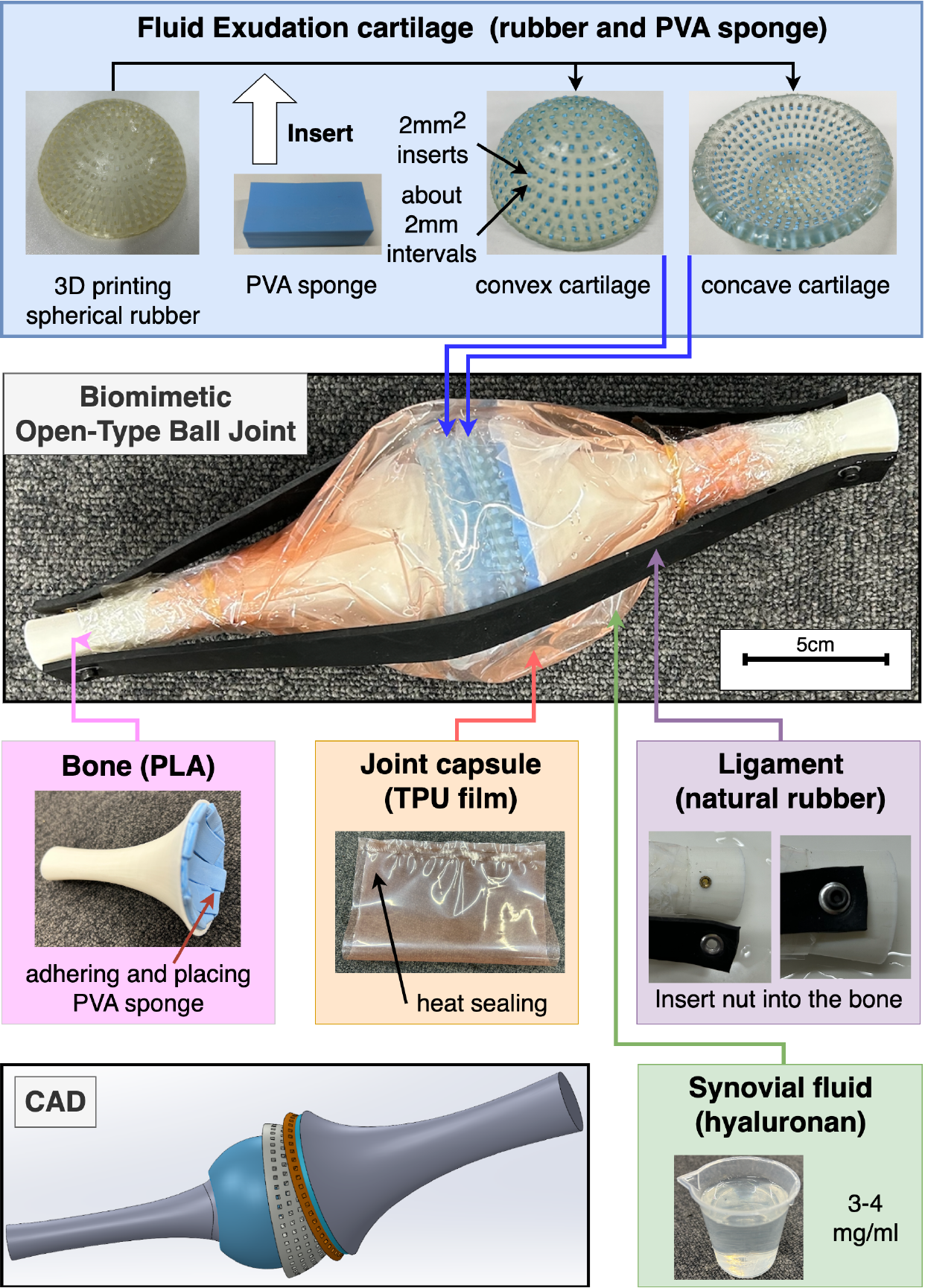}
    \caption{Figure illustrating the construction method of an open-type ball joint utilizing fluid exudation cartilage.}
    \label{fig:construction-open-type-ball-joint}
  \end{center}
\end{figure}

\begin{figure}[t]
  \begin{center}
    \includegraphics[width=\linewidth]{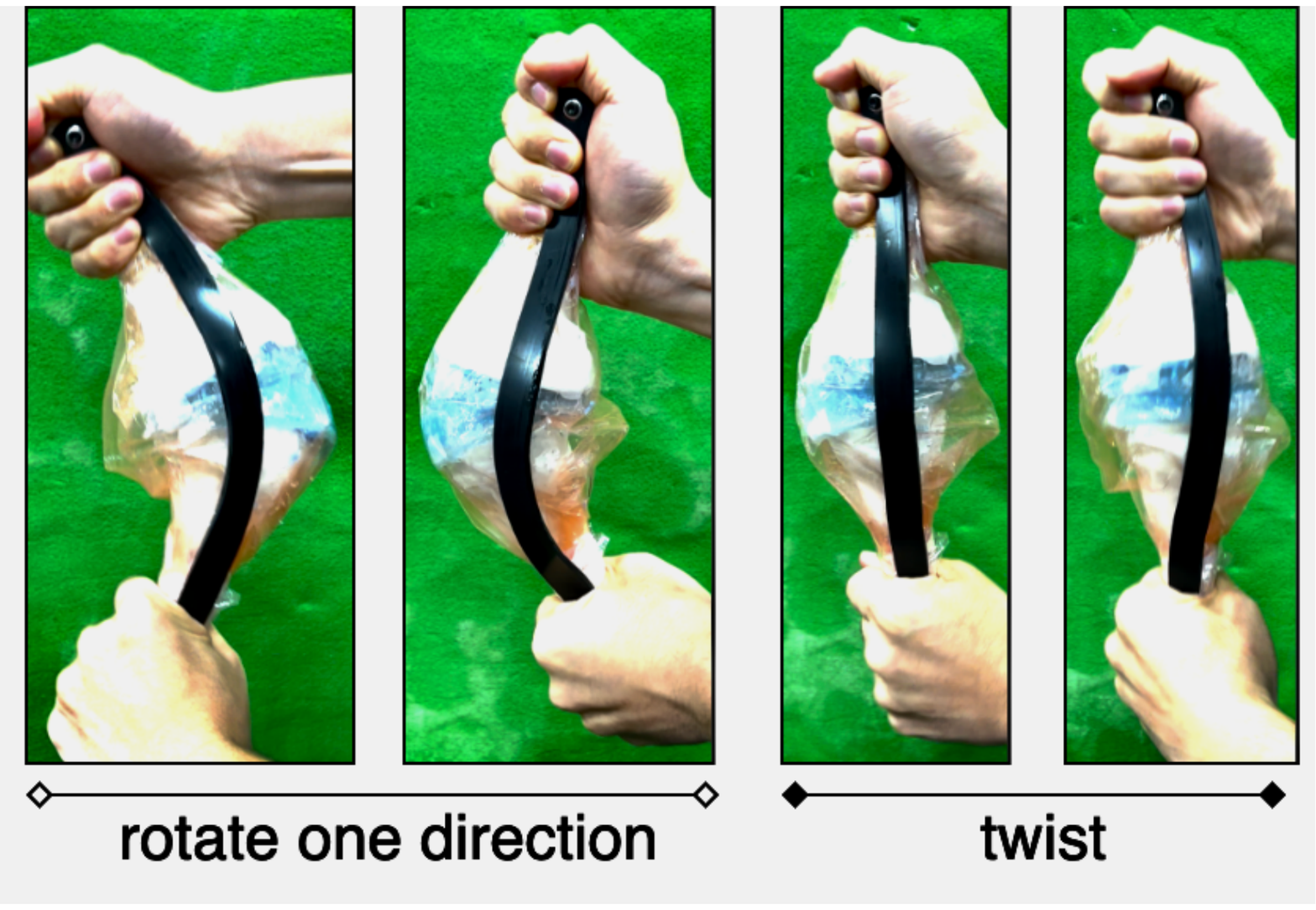}
    \caption{Figure illustrating the manual movement of an open-type ball joint.}
    \label{fig:range-of-motion}
  \end{center}
\end{figure}

}%

\section{Conclusion}
\label{sec:conclusion}
\switchlanguage
{%
  In this study, our goal was to replicate the low-friction nature of human joints by emulating fluid-exuding cartilage.
  The cartilage sheet, made of resin rubber with PVA sponge inserts, mimics human cartilage.
  Under applied load, the sheet deforms, compressing the sponge and promoting the exudation of synovial fluid.
  This fluid reduces friction through lubrication as it is squeezed towards the articulating bones and cartilage surfaces.
  Upon load reduction, the rubber and sponge recover, replenishing the exuded fluid from the joint capsule back to the sponge, making both functions reusable.
  The contribution of the fabricated fluid-exuding cartilage to low friction was confirmed through friction coefficient measurement tests.

  Furthermore, we attached the cartilage sheet to bones and constructed an open-type ball joint along with ligaments, synovial fluid, and the joint capsule.
  It was verified that the cartilage sheet proposed in this study could be 3D-printed in actual size and in a curved shape, enabling the construction of an open-type ball joint.

  The liquid exuding cartilage created in this study could serve as a basic element for creating open-type ball joints, similar to those in the human body, that contain liquid.
  This opens up new directions for bio-inspired robots that have not dealt much with liquids before.
  Additionally, human body's open-type ball joints have the characteristic of having both low friction and shock absorption, as well as possessing multiple drive axes, not just one.
  As research progresses towards practicality, applications are expected in general robotics as well.
  Moreover, discussions on systems that include soft elements such as bodies and cartilage using liquid, regarding how to sense and control them, may make us understand the human from a constructivistic viewpoint.

  The experiments in this study were conducted with a limited number of samples, necessitating the evaluation of dispersion through repeated experiments to reduce uncertainty and noise.
  Further evaluations should include different rubber elastic bodies, and friction coefficient measurements for spherical cartilage should be conducted, alongside the development of measurement devices.

  Future research should also advance studies on ligaments and joint capsules to improve properties such as the range of motion, mimicking biological joints.
  Additionally, we plan to attach motors to the bones of the constructed open-type ball joints to verify its characteristics, including impact resistance, softness, and adaptability to the environment during movement, essential for understanding the joint's behavior in practical applications.

}%
{%
本研究では，人間の軟骨を参考にして，人体関節の低摩擦性を実現する液体滲出軟骨を模倣再現した．
人体軟骨を模倣再現した軟骨シートは，レジンゴムにPVAスポンジを挿入することで構成される．
荷重量に応じて変形し，その変形量に応じてスポンジが押されることで，スポンジに蓄えられていた滑液が滲出する．
その結果，荷重がかかることで，強く接する関節内の骨や軟骨の間，
\textcolor{red}{
つまり荷重面へと流体がおしだされる．
おしだされた流体は
}
潤滑現象を引き起こし，摩擦を減少させる．
荷重がかからなくなるにつれてゴムおよびスポンジの変形が戻ることで，滲出した液は関節包内に満ちている滑液からスポンジへと再補充される．
その結果，液体滲出機能およびそれによる潤滑機能は反復利用することができる．
実際に作成した液体滲出軟骨が低摩擦性に貢献することは摩擦係数測定試験を行うことで確認できた．
さらに，軟骨シートを実際に模擬骨にとりつけ，靭帯，関節液，関節包とともに開放型球関節の構成も行った．
本研究で提案した軟骨シートが，等身大かつ曲面状に3Dプリンティングでき，開放型球関節を構成できることまで確認できた．

\textcolor{red}{
本研究で作成した液体滲出軟骨は，人間の身体と同じような，液体を内包した開放型球関節を作成するための基礎要素となりうる．
液体をあまり扱ってこなかった従来の生体模倣ロボットに新たな方向性を開くものである．
また，人間の身体の開放型球関節は，低摩擦性と衝撃吸収性を兼ね備えて，一つの駆動軸だけでなく複数の駆動軸を併せ持つ特徴を持つ．
実用性に至るまで研究が進めば，一般的なロボティクスにおいても応用が見込まれるものである．
さらに，液体を用いた身体や軟骨など柔らかな要素を含む身体を，どのようにセンシングしてどのように制御するかといったシステムの議論は，構成論的な人体理解への可能性も含んでいる．
}

\textcolor{red}{
ただし，本研究の実験は限られたサンプルで行われた．
実験の操作者による不確実性やノイズを消すためにも，繰り返し実験による分散の評価を行う必要がある．
また，異なるゴム弾性体における評価も行うべきである．
さらには，実際のロボットへの応用を見越して，平面軟骨シートだけではなく，構成した球状軟骨についても摩擦係数評価を測定装置の開発も含めて行っていくべきである．
上記のような基礎的評価を続けていく必要がある．
}

\textcolor{red}{
その上で
}
今後の研究では，
\textcolor{red}{
軟骨だけでなく，靭帯や関節包に関しても研究を進めて，可動域をはじめとした様々な性質を生体関節に倣って向上させていく必要がある．
加えて，
}
実際に作成した開放型球関節を構成する骨に筋肉の役割を果たすモーターを取り付けて駆動させ，関節全体としての特性を検証したいと考えている．
特に，耐衝撃性・柔らかさ・環境への馴染みやすさといった点について，関節を動かしながら詳しく検証していく予定である．
}%

{
  \bibliographystyle{IEEEtran}
  \bibliography{bib}
}

\end{document}